\pdfoutput=1
\documentclass[11pt]{article}

\usepackage[final]{ACL2024}
\usepackage{graphicx}
\usepackage{times}
\usepackage{latexsym}
\usepackage{amsmath}
\usepackage{multirow}
\usepackage{array}
\usepackage{comment}
\usepackage[framemethod=TikZ]{mdframed}
\newcolumntype{C}[1]{>{\centering\arraybackslash}p{#1}}
\newcolumntype{L}[1]{>{\raggedright\arraybackslash}p{#1}}
\usepackage[T1]{fontenc}
\usepackage[utf8]{inputenc}

\usepackage{microtype}

\usepackage{inconsolata}
\usepackage{booktabs}
\usepackage{amsmath}
\usepackage{amssymb}

\newcommand{\eg}{\textit{e.g.}}
\newcommand{\ie}{\textit{i.e.}}

\title{Small Models, Big Insights: Leveraging Slim Proxy Models to Decide When and What to Retrieve for LLMs}

\author{Jiejun Tan$^{1}$\thanks{This work was done when Jiejun Tan was doing internship at Baichuan Intelligent Technology.}, Zhicheng Dou$^{1}$\thanks{Corresponding author.}, Yutao Zhu$^1$, Peidong Guo$^2$
\\ \textbf{Kun Fang}$^2$, \and \textbf{Ji-Rong Wen}$^1$ \\
$^1$Gaoling School of Artificial Intelligence, Renmin University of China \\
$^2$ Baichuan Intelligent Technology \\
\texttt{\{zstanjj, dou\}@ruc.edu.cn}
}

\begin{document}
\maketitle
\def\thefootnote{\arabic{footnote}}

\begin{abstract}
% TLDR
%SlimPLM improves LLMs' question-answering performance by using a slim proxy model to identify and retrieve only the missing knowledge, significantly reducing computational costs. 
The integration of large language models (LLMs) and search engines represents a significant evolution in knowledge acquisition methodologies. However, determining the knowledge that an LLM already possesses and the knowledge that requires the help of a search engine remains an unresolved issue. Most existing methods solve this problem through the results of preliminary answers or reasoning done by the LLM itself, but this incurs excessively high computational costs.
This paper introduces a novel collaborative approach, namely SlimPLM, that detects missing knowledge in LLMs with a slim proxy model, to enhance the LLM's knowledge acquisition process. We employ a proxy model which has far fewer parameters, and take its answers as heuristic answers. Heuristic answers are then utilized to predict the knowledge required to answer the user question, as well as the known and unknown knowledge within the LLM. We only conduct retrieval for the missing knowledge in questions that the LLM does not know.
Extensive experimental results on five datasets with two LLMs demonstrate a notable improvement in the end-to-end performance of LLMs in question-answering tasks, achieving or surpassing current state-of-the-art models with lower LLM inference costs.\footnote{Our code and datasets are available at \url{https://github.com/plageon/SlimPlm}.} %Experiments also shows our method maintains system efficiency, avoiding extra load and brought by LLM inferences.
% In summary, our work shows answers from a proxy model can can inspire retrieval augmented generation in LLMs, presenting a new paradigm in retrieval augmented generation. This approach not only enhances the accuracy of knowledge determination in LLM responses but also offers a targeted and selective search process, substantially elevating the overall efficacy of LLMs in knowledge question-answering tasks.
% LLMs excel in reasoning and commonsense knowledge, yet they falter with non-commonsense or novel information. Conversely, search engines retrieve up-to-date and niche knowledge effectively but lack reasoning and summarizing skills. 
\end{abstract}

\section{Introduction}
Large language models (LLMs) have demonstrated significant prowess in various natural language processing (NLP) tasks~\cite{Achiam2023GPT4TR}, attributed to their advanced language comprehension and generation capabilities. Despite being trained on extensive text corpora, these models occasionally produce hallucinated content~\cite{Zhou2020DetectingHC,Maynez2020OnFA}. To tackle this problem, the integration of retrieval systems with LLMs has been proposed, enabling access to external knowledge bases for more accurate and reliable text generation.

% Generative LLMs have already shown powerful capabilities and immense potential for applications~\cite{Achiam2023GPT4TR}. 
% The combination of search engines and LLMs has changed our existing methods of knowledge acquisition. LLMs possess reasoning abilities and much commonsense knowledge~\cite{Petroni2019LanguageMA,Jiang2019HowCW,Roberts2020HowMK}, but they struggle with hallucination when dealing with non-commonsense or new knowledge~\cite{Zhou2020DetectingHC,Maynez2020OnFA}. Search engines can retrieve the latest, and specialized domain knowledge, but lack the capabilities for reasoning and summarizing. 

Retrieval-augmented generation (RAG) involves using a retrieval system to supplement LLMs with relevant external information, thereby improving text generation quality~\citep{Peng2023CheckYF,He2022RethinkingWR}. 
Yet, recent studies have suggested that retrieval may not always be beneficial. In cases where LLMs can adequately respond without external knowledge, retrieval may introduce irrelevant information, potentially degrading performance~\cite{Kadavath2022LanguageM,Wang2023SelfKnowledgeGR,Shi2023LargeLM,Petroni2020HowCA}.
% However, it is not the best approach to perform search enhancement for LLMs under all tasks. Existing work has proven that LLMs are vulnerable to irrelevant contexts, especially for questions that the LLM can independently answer well enough~\cite{Kadavath2022LanguageM,Wang2023SelfKnowledgeGR,Shi2023LargeLM,Petroni2020HowCA}. 
Therefore, it is critical to determine when retrieval is necessary for user questions~\cite{Shuster2021RetrievalAR}.
% a retrieval augmented LLM can better play to its strengths and avoid weaknesses by performing searches only on user questions where the LLM needs to retrieve information.
The challenge lies in identifying questions that exceed the LLMs' intrinsic knowledge and require external retrieval, due to the prevalence of content hallucination. Efforts to address this challenge can be categorized into two groups:
(1) The first group of methods involves fine-tuning LLMs for RAG scenarios, allowing them to autonomously signal the need for external knowledge~\cite{Nakano2021WebGPTBQ,Liu2023WebGLMTA,Qin2023ToolLW}. 
% For example, some methods enable LLMs to generate special tokens to indicate the need for external knowledge assistance~\cite{Nakano2021WebGPTBQ,Liu2023WebGLMTA,Qin2023ToolLW}. 
This method, while effective, demands substantial computational resources and risks diminishing the LLMs' general capabilities due to potential catastrophic forgetting~\cite{Kotha2023UnderstandingCF,Zhai2023InvestigatingTC}. 
(2) The second category avoids direct tuning of LLMs, assessing the necessity for retrieval based on the quality of the generated content or specific indicators within it~\cite{Ram2023InContextRL,Min2022RethinkingTR}. However, this approach still has its drawbacks, as it requires multiple inferences, thereby increasing both the inference costs and the latency of responses to user questions.

In light of this, we put forward a question: \textit{Is it feasible to employ a proxy model with a relatively smaller parameter size to facilitate effective retrieval results for an LLM?} Theoretically, existing decoder-only language models share similar Transformer structures, and they are pre-trained on some common text corpora, such as Common Crawl web pages, books, and Wikipedia pages~\citep{Touvron2023Llama2O,Bai2023QwenTR,Scao2022BLOOMA1,Almazrouei2023TheFS,Zhang2024TinyLlamaAO}. Therefore, it is possible for them to reach a consensus on relative mastery over different knowledge and the necessity of retrieval. Our preliminary quantitative analysis, shown in Section~\ref{knowledge-gap}, also supports this hypothesis. 
% The experimental results indicate that models of different sizes show similar proficiency on easy-to-understand topics, but show large performance differences on more difficult problems. This further validates the possibility of employing a proxy model to help determine the necessity of retrieval.
The experimental results show that on questions well understood by the LLM, the relatively smaller language model also has considerable knowledge. The gap between larger and smaller LLMs mainly manifests in questions they do not understand. This further validates the possibility of employing a proxy model to help determine the necessity of retrieval.
% We rely on this answer to gauge the larger model's understanding of knowledge pertinent to users' questions. This method is based on two assumptions regarding two language models with different parameter scales: (1) Despite their gap in ability, both language models show a remarkable consensus on relative mastery over different knowledge; (2) Their proficiency levels are closely matched on well-understood topics, but diverge significantly on less familiar ones.

Based on our analysis, in this paper, we introduce a novel approach, called \textbf{SlimPLM} (\textbf{Slim} \textbf{P}roxy \textbf{L}anguage \textbf{M}odel), which leverages a relatively smaller language model as a ``proxy model'' to help determine when and how to perform retrieval for LLMs. Specifically, for a user question, SlimPLM first uses the proxy model to generate a preliminary ``heuristic answer''. This heuristic answer serves two purposes. First, it is evaluated by a lightweight model designed to assess the necessity for retrieval. If this evaluation shows that the heuristic answer is of high quality, it implies that the question may be addressed directly by LLMs without additional information retrieval. In contrast, a lower-quality answer triggers the retrieval process to identify and supplement missing knowledge. To facilitate this, SlimPLM utilizes the heuristic answer again to generate multiple queries, each reflecting a specific aspect of the initial response. These queries are then individually assessed for their need for retrieval, filtering out queries that do not require retrieval. By this means, the remaining queries can retrieve more relevant knowledge that is lacking in LLMs. The integration of SlimPLM into existing RAG frameworks offers a flexible and effective enhancement without notably increasing computational costs or response latency. Experimental results across five commonly used question-answering datasets validate SlimPLM's effectiveness in determining the necessity for retrieval and improving retrieval results.

Our contributions are threefold:
(1) We propose a novel approach that leverages a small proxy model to generate heuristic answers, helping determine when and how to perform retrieval for LLMs.
(2) We devise a retrieval necessity judgment model based on the heuristic answer. It is capable of accurately identifying which queries necessitate further information retrieval.
(3) We formulate a query rewriting strategy that decomposes the heuristic answer into distinct claims. This is complemented by a claim-based filtering mechanism to enhance the relevance of the retrieval results for LLMs' text generation.

% (3) We design a query rewriting method based on the decomposition of the heuristic answer and a claim-based filter to refine these queries, which provide more relevant retrieval results for LLMs' generation. 

% We introduce a novel approach, SlimPLM, for RAG. With only a minor increase in computational cost, it can accurately distinguish between the knowledge that is missing in the LLMs, and conduct effective retrieval. 

% (2) Our new approach for search enhancement improves the end-to-end effectiveness of LLMs in knowledge question-answering. We achieved state-of-the-art performance while having fewer LLM inference times.

% (3) We investigate the knowledge capabilities of language models with varying parameters trained on similar corpora. Our findings reveal an overlap in their knowledge capabilities, providing theoretical support for our approach.

\begin{figure*}
    \centering
    \includegraphics[width=1\textwidth]{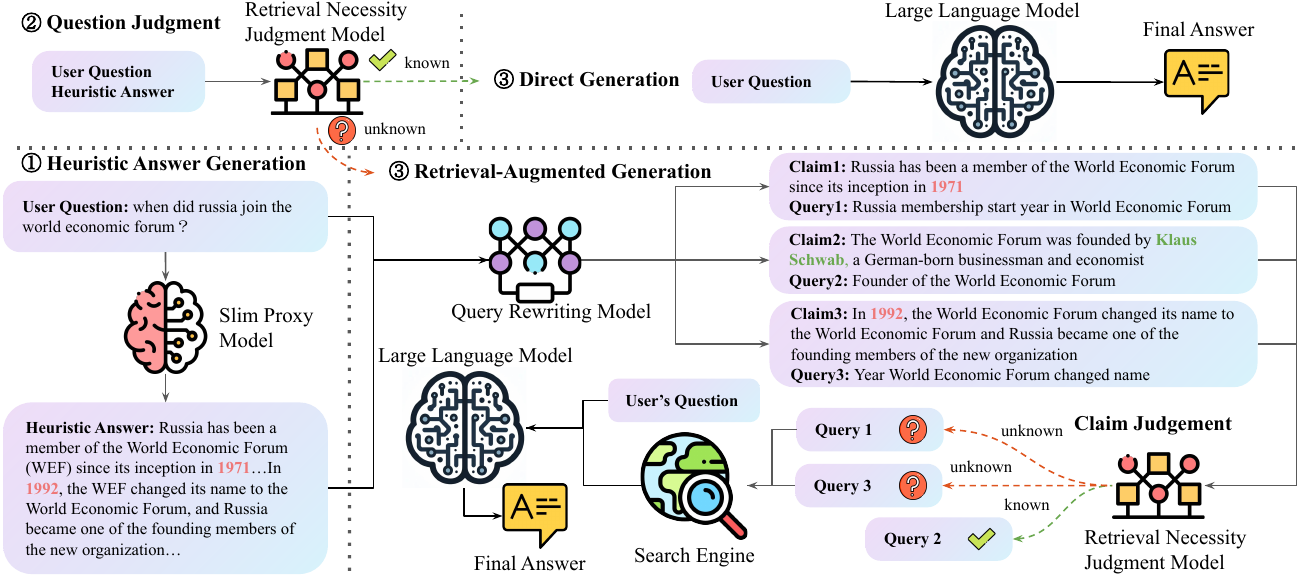}
    \caption{A display of the main process of SlimPLM. Solid lines with arrows represent the flow of data, while dashed lines with arrows signify control signals from the retrieval necessity judgment model. Step 1 and step 2 are mandatory in the pipeline, but step 3 involves choosing between direct generation and RAG.}
    \label{main-pipeline}
\end{figure*}

\section{Related Work}
\subsection{Retrieval-Augmented Generation (RAG)}
RAG has been studied for a long time. In the era of pre-trained language models, RAG has been applied to provide models with relevant knowledge, significantly enhancing the generation quality in applications such as dialogue systems~\cite{DBLP:conf/sigir/TahamiGS20,DBLP:conf/wsdm/TaoWXHZY19} and question-answering systems~\cite{DBLP:conf/eacl/IzacardG21,DBLP:conf/sigir/TahamiGS20}. With the development of LLMs, RAG has emerged as a crucial strategy to tackle the problem of hallucination and outdated information~\cite{Shuster2021RetrievalAR,White2023NavigatingCS}.

% Before the rise of LLMs, small language models~\cite{Raffel2019ExploringTL,Lewis2019BARTDS} possess limited knowledge, so retrieval-augmentation is used to assist knowledge-incentive QA tasks. Previous works often design a fixed module for retrieval-augmentation~\cite{Sun2022RecitationAugmentedLM,Lewis2020RetrievalAugmentedGF,Izacard2020LeveragingPR}.
% Since the release of ChatGPT, LLMs have been developing rapidly~\cite{Zhao2023ASO}. LLMs possess abundant intrinsic knowledge, but hallucination issue means they also have to rely on retrieval-augmentation to improve factuality~\cite{Shuster2021RetrievalAR,White2023NavigatingCS}.

The mainstream RAG methods follow a ``retrieve-then-read'' architecture. In this setup, a retrieval module first gathers external knowledge, providing additional context that is subsequently processed by LLMs to generate the final output~\cite{Ram2023InContextRL,Yu2023ImprovingLM}. Typically, a RAG pipeline~\cite{Zhu2023LargeLM,Liu2023RETALLMAR,Shi2023REPLUGRB} includes several components: a query rewriter that refines the initial query~\cite{Wang2023Query2docQE,Gao2022PreciseZD}, a retriever that fetches relevant documents~\cite{Guu2020REALMRL,Neelakantan2022TextAC}, a filter or reranker~\cite{Yoran2023MakingRL,Yu2023ChainofNoteER,Xu2023RECOMPIR} that ensures only the most relevant knowledge is kept, and an LLM as reader that generates the final results. To optimize these systems, some approaches focus on enhancing individual components of the RAG architecture to improve overall performance~\cite{inters,bider}, while others involve direct fine-tuning of the LLM to better integrate with RAG-specific tasks~\cite{Asai2023SelfRAGLT,Kadavath2022LanguageM}.
% For example, GAR~\cite{Mao2020GenerationAugmentedRF} focuses on augmenting queries through text generation of relevant contexts without relying on external resources for supervision. REALM~\cite{Guu2020REALMRL} pre-trains the knowledge retriever in an unsupervised manner. 
% This is achieved using masked language modeling as the learning signal and allowing backpropagation through a search step that considers millions of documents. 

\subsection{Retrieval Necessity Judgment}
In a Retrieval-Augmented Generation (RAG) system, a critical challenge is determining when to initiate the retrieval process. Several approaches have been proposed to address this issue:

(1) Fine-tuning Large Language Models (LLMs) has proven effective but comes with substantial computational costs~\cite{Qin2023WebCPMIW,Lin2022TeachingMT}. Some studies have focused on fine-tuning the LLM to mimic human-like web browsing behavior~\cite{Schick2023ToolformerLM,Nakano2021WebGPTBQ}. Self-RAG~\cite{Asai2023SelfRAGLT} introduces special tokens known as reflection tokens to regulate retrieval behavior.

(2) Another intuitive approach involves evaluating the LLM's confidence based on the logits generated by the model~\cite{Jiang2020HowCW,Guo2017OnCO}. FLARE~\cite{Jiang2023ActiveRA} dynamically activates RAG if the logits fall below a predefined threshold.

(3) Other research has employed iterative prompting to determine if additional information is required~\cite{Wei2022ChainOT,Liu2021WhatMG,Rubin2021LearningTR}, or has combined Chain-of-Thought prompting~\cite{Wei2022ChainOT} with RAG~\cite{Press2022MeasuringAN,Khattab2022DemonstrateSearchPredictCR}. For instance, ReAct~\cite{Yao2022ReActSR} alternates between generating thoughts and actions, creating a sequence of thought-action-observation steps.

(4) Evaluating the complexity or popularity of user questions to assess the need for retrieval is also a feasible approach~\cite{Mallen2022WhenNT}. SKR~\cite{Wang2023SelfKnowledgeGR} refers to similar questions it has previously encountered to determine the necessity of retrieval.

Distinct from these existing methods, SlimPLM evaluates the necessity of retrieval by analyzing the answer generated by a smaller LLM. This approach does not increase LLM inference times while enhancing judgment accuracy.

\subsection{Query Formulation}
In addition to determining when to retrieve information, the question of what to retrieve is also of great importance. A simplistic approach that merely judges the necessity of retrieval based on the user's query would be inadequate. The aforementioned retrieval necessity judgment work has also proposed solutions for query rewriting. Numerous studies have encouraged the use of Large Language Models (LLMs) to autonomously generate queries~\cite{Yao2022ReActSR,Press2022MeasuringAN,Schick2023ToolformerLM}. Some research has utilized previously generated content as queries~\cite{Shao2023EnhancingRL,Asai2023SelfRAGLT}, or have taken a further step by masking low logit tokens within the generated content~\cite{Jiang2023ActiveRA}. Other studies have employed specially fine-tuned query rewriting models to rewrite either the user's question or previously generated content~\cite{Wang2023Query2docQE,Ma2023QueryRF}.

In contrast, SlimPLM formulates queries by meticulously analyzing the answers generated by a smaller LLM, thereby providing an accurate understanding of the required knowledge.

% Nevertheless, mere analysis for user question overlooks the knowledge capabilities of LLMs. Our method, on the other hand, keeps the one-off generation while takes the knowledge capabilities of LLMs into account. We pre-determine the knowledge required to answer the current question, identifying both the knowledge already possessed by the large model and the knowledge that the large model lacks.

% Our approach is to introduce a smaller, lighter language model with fewer parameters than the LLM to provide heuristic answers. Also, we introduce a model for analyzing the knowledge in heuristic answers and another for query rewriting based on these heuristic answers. This way, we can choose whether or not to provide reference documents without extra calls to the LLM, and the reference documents provided will contain only the knowledge needed by the LLM.

\section{Methodology}
In this paper, we aim to leverage a relatively smaller model as the \textit{proxy model} to determine whether the user-issued question requires supplementary retrieval results and further provide clues for retrieving relevant knowledge. Our method, SlimPLM, can be flexibly used as a plug-in to various retrieval-augmented generation scenarios, without additional training requirements. The illustration of our method is shown in Figure~\ref{main-pipeline}.
% We propose SlimPLM, a pluggable component for retrieval-augmented generation. The basic idea of our method is leveraging 
% This method does not require additional training for LLMs, but can effectively enhance the end-to-end performance of LLMs. We formally define a language model with a smaller parameter scale than the LLM as the \textit{proxy model}, and its direct answer to the user question as a heuristic answer. Our major pipeline is demonstrated in Figure~\ref{main-pipeline}.

% \subsection{Knowledge boundaries for language models}

% Whether it's proxy models or LLMs, they do not have clear knowledge boundaries, and all the knowledge possessed by the models is a matter of probability. From their most confident common knowledge to obscure knowledge, it is a process of give and take from correct facts to hallucinations. Some uncertain knowledge in large models might be correctly answered under the guidance of certain prompts, while it might not be correctly answered under other prompts. For these vague pieces of knowledge, it is also best to provide search results.

\subsection{Problem Formulation \& Framework}
Before diving into the details of our method, we first formulate the concept and notations involved in this paper.

Given a user input $x$ and a text corpus (\eg, a Wikipedia dump) $D=\{d_i\}_{i=1}^{N}$ of size $N$, models are expected to generate the annotated answer $y$. 
To obtain the information in $D$ that is relevant to $x$, a retriever (R) is employed. This retriever takes a query $q$ as input and returns a relevant text list $D_{\text{ref}}=R(q)$. Typically, the user input $x$ is used as the query, namely $q=x$. However, existing studies have demonstrated that using refined queries for retrieval can improve the final generation quality~\cite{Gao2022PreciseZD,Wang2023Query2docQE}. Therefore, we denote the refined queries as $\{q_1, \ldots, q_n\}$. 
% These refined queries are often obtained by specific query rewriting models~\cite{}. 
With these refined queries, a collection of relevant retrieval results $D_{\text{ref}} = {R(q_1), \ldots, R(q_n)}$ is assembled to support the generation process, formalized as $\hat{y} = \text{LLM}(D_{\text{ref}}, x)$.
% With the refined queries, we can get a set of retrieval results for subsequent generation, namely $ D_{\text{ref}} = \{R(q_1), \ldots, R(q_n)\}$. After obtaining retrieval results, the answer generation process can be defined as: $ \hat{y} = \text{LLM}(D_{\text{ref}}, x) $. 
Note that, when $D_{\text{ref}}=\varnothing$, the process degenerates to normal generation without retrieval.
% and the reference documents are directly retrieved based on the user's input, denoted as $D_{\text{ref}} = R(x)$. 
% If query rewriting is involved in the RAG pipeline, then the query rewriting model can generate multiple queries $ \{q_1, \ldots, q_n\} $ from the user input, and obtain reference documents for each, which are combined to serve as the final reference document set, $ D_{\text{ref}} = D_{Q} = \{R(q_1), \ldots, R(q_n)\} $.

We define a proxy model (PM), which is implemented by a relatively smaller LLM. The proxy model generates an answer for the input $x$ as: 
\begin{align}
    \hat{a} = \text{PM}(x), \label{eq:a}
\end{align}
where $\hat{a}$ is called a \textit{heuristic answer} in this paper. This heuristic answer serves two purposes: (1) It is used for determining whether the retrieval is necessary for the current input $x$. The determination is made by a retrieval necessity judgment model (introduced in Section~\ref{retrieval-necessity-judgement}). (2) It also provides clues for query rewriting. The query rewriting results will help identify knowledge gaps within the LLM that necessitate further retrieval (introduced in Section~\ref{retrieval-target-determination}).
% which knowledge is lacked by the LLM and requires additional retrieval . 

\subsection{Retrieval Necessity Judgment}
\label{retrieval-necessity-judgement}

Because existing LLMs are typically trained on common corpora (such as CommonCrawl and Wikipedia~\cite{Touvron2023Llama2O,Penedo2023TheRD,Gao2020ThePA}) and employ a similar Transformer decoder-based architecture, it is promising to leverage a smaller LLM for judging the knowledge mastered by larger LLMs and determining the need for additional retrieval. Thus, we propose a retrieval necessity judgment component. 

% We aim to evaluate the retrieval necessity of a user question through a retrieval necessity judgment model $\text{NJ}$. However, making a general judgment on the necessity of retrieval for user questions is insufficient, as the large language model's mastery over different aspects of a question is inconsistent. So we further conduct retrieval necessity judgment on each rewritten query using NJ. The heuristic answers from the proxy model are a crucial basis for judging retrieval necessity. 

% \subsubsection{Retrieval Necessity Label Collection}

\paragraph{Judgment Model}
Given the heuristic answer $\hat{a}$ generated by the proxy model (Equation~\ref{eq:a}), we fine-tune a judgment model RJ (implemented by Llama2-7B in our experiments) by using both the user input $x$ and the heuristic answer $\hat{a}$. We use the following instructions for fine-tuning:

\begin{mdframed}[backgroundcolor=gray!5, roundcorner=3pt, innerleftmargin=10pt, innerrightmargin=10pt, innertopmargin=10pt, innerbottommargin=10pt]
\small
\texttt{Input:}

\noindent\texttt{\textcolor{orange}{<SYS> You are a helpful assistant. Your task is to parse user input into structured formats and accomplish the task according to the heuristic answer. </SYS>}}

\noindent\texttt{\textcolor{orange}{Heuristic answer:} \textcolor{cyan}{\textbf{\{Heuristic Answer\}}}}

\noindent\texttt{\textcolor{orange}{Question:} \textcolor{cyan}{\textbf{\{user question\}}}}

\noindent\texttt{\textcolor{orange}{Retrieval Necessity Judgment Output:}}

\noindent\texttt{Output:}

\noindent\texttt{\textcolor{teal}{\textbf{Known (True / False)}}}

\end{mdframed}

\noindent After fine-tuning, the RJ model can predict whether a user question needs further retrieval with the help of the heuristic answer.

\paragraph{Judgment Label Collection} 
To fine-tune the RJ model, we need to collect training samples with reliable labels. Existing studies~\cite{Wang2023SelfKnowledgeGR} have proposed an annotation strategy that compares the models' outputs generated with and without retrieval. In our preliminary study, we find that this strategy is highly influenced by the capability of the retriever and the completeness of the corpus, leading to annotations that cannot accurately reflect the model's necessity for search. To tackle this problem, we propose to leverage the quality of our heuristic answers, \ie, if the quality of the heuristic answer is higher than a predefined threshold, we infer that the question can be well answered without retrieval; otherwise, we consider retrieval necessary. 

Specifically, we collect samples with short answers from existing question-answering datasets and employ the matching ratio between the heuristic answers and the ground-truth answers as the metric. Compared to rouge scores~\cite{Lin2004ROUGEAP} or perplexity~\cite{chip2019evaluation}, this metric can better align with the evaluation and reflect the generation quality. 
% We assume that on knowledge QA tasks, a matching ratio of short answers best reflects the output quality of language models, compared to rouge score~\cite{Lin2004ROUGEAP} or perplexity~\cite{chip2019evaluation}.
Notably, while we only use short answers for label collection, the obtained model can well generalize to different datasets, such as long-form QA datasets.
% We fine-tune the retrieval necessity judgment model on data with matching ratios of short answers. The model's learned ability to judge retrieval necessity can be generalized to different datasets, even if they have different annotation formats. Moreover, it can also generalize to the input form of a single claim. The instructions we designed and the model output are displayed as follows.
Formally, for a question with multiple short answers $ Y = \{y_1, y_2, \ldots, y_n\} $, we compute the matching ratio $r$ between $\hat{a}$ and $Y$ as: 
\begin{align}
    r = \frac{|\{y \mid y \in \hat{a}\wedge y \in Y \}|}{|Y|}.
\end{align}
Then, we set a threshold $ \theta $ and obtain the label as:
\begin{equation*}
    \text{Label}(\hat{a}, x) = 
    \begin{cases} 
    \text{Known (True)}, & \text{if } r > \theta; \\
    \text{Known (False)}, & \text{otherwise.}
    \end{cases}
\end{equation*}

\subsection{Retrieval Target Determination}
\label{retrieval-target-determination}
After determining the necessity of retrieval, the next question is how to perform effective retrieval. A straightforward method is using user input $x$ as the query to retrieve relevant information from the corpora $D$. However, many studies have reported that the information retrieved by $x$ may lose details and introduce redundant content
~\cite{Wang2023Query2docQE}. To address this issue, we propose a query rewrite method based on the heuristic answers and a query filter method to refine these rewritten queries.

\paragraph{Heuristic Answer-Driven Query Rewrite}
Restricted by parameter scale, the proxy model often hallucinates during the process of answering questions, but the direction in which they answer questions is heuristic~\cite{Dhuliawala2023ChainofVerificationRH,Gao2022PreciseZD}. They can extend related aspects and sub-topics of thought when analyzing questions. Inspired by claim decomposition operation intended for factual evaluation~\cite{Min2023FActScoreFA,Kryscinski2019EvaluatingTF}, we perform query rewriting based on each fact mentioned in the heuristic answer given by the proxy models. The specific operations are as follows: we decompose the heuristic answer $ \hat{a} $ into multiple claims related to the question, $ \{c_1, c_2, \ldots, c_n\} $, where each claim related to the question can lead to a query, $ \{q_{c_1}, q_{c_2}, \ldots, q_{c_n}\} $. In addition, we combine the query rewrites directly derived from the user's input $ \{q_{x_1}, q_{x_2}, \ldots, q_{x_n}\} $. Our query rewriting model QR takes the user question and the heuristic answer as input and outputs all query rewrite results, $ \text{QR}(x, \hat{a}) = \{q_{x_1}, \ldots, q_{x_n}, q_{c_1}, \ldots, q_{c_n}\} $.

To train the query rewriting model, we collect and annotate a dataset with the help of GPT-4\cite{Achiam2023GPT4TR}. In each dataset used in our experiments, we sample 1,000 user questions. We utilize the method of instruction fine-tuning~\cite{Ouyang2022TrainingLM,Chung2022ScalingIL} to fine-tune a decoder-only generative model, accomplishing the task of claim extraction and query rewriting in a single round. Our instructions and the model output are displayed as follows.

\begin{mdframed}[backgroundcolor=gray!5, roundcorner=3pt, innerleftmargin=10pt, innerrightmargin=10pt, innertopmargin=10pt, innerbottommargin=10pt]
\small
\texttt{Input:}

\noindent\texttt{\textcolor{orange}{<SYS> You are a helpful assistant. Your task is to parse user input into structured formats and accomplish the task according to the heuristic answer. </SYS>}}

\noindent\texttt{\textcolor{orange}{Heuristic answer:}} \texttt{\textcolor{cyan}{\textbf{\{Heuristic Answer\}}}}

\noindent\texttt{\textcolor{orange}{Question:}} 
\texttt{\textcolor{cyan}{\textbf{\{User Question\}}}}

\noindent\texttt{\textcolor{orange}{Query Rewrite Output:}}

\noindent\texttt{Output:}

\noindent\texttt{\textcolor{teal}{\textbf{<Claim> Claim 1 <Query> Query 1 <Claim> Claim 2 <Query> Query 2, ...}}}

% <Questions> <Question(\textbf{Question})> <Query(\textbf{Query})> <NeedSearch(\textbf{True/False})> ... </Questions> 
% <Claims> <Claim(\textbf{Claim})> <NeedSearch(\textbf{True/False})> <Query(\textbf{Query})> ... </Claims>

\end{mdframed}

\paragraph{Claim-based Query Filter}
% Retrieval necessity judgment is separated into two stages: question-level retrieval necessity judgment and claim-level retrieval necessity judgment. Question-level retrieval necessity judgment decides whether to use RAG or direct generation. Claim-level retrieval necessity judgment decides which claims requires retrieval.
In the previous step, our method generates several rewritten queries QR$(x, \hat{a})$, which correspond to the claims in the heuristic answers. 

To achieve this, we reuse the judgment model RJ trained in Section~\ref{retrieval-necessity-judgement}. Specifically, we replace the input of the heuristic answer by the extracted claim and the input of user questions by the rewritten query. Then, the model can predict whether the rewritten query requires external knowledge from retrieval. We only perform retrieval when the result is Known (False), namely, we have:
\begin{align*}
    D_{\text{ref}} = \{R(q_{c_i}) | \text{RJ}(c_i, q_{c_i}) = \text{Known (False)}\}. 
\end{align*}
By this means, we can obtain the retrieved result set $ D_{\text{ref}}$ that only contains the knowledge missing by the LLM.

% \begin{equation*}
%     \hat{y} = 
%     \begin{cases} 
%     \text{LM}(x), & \text{if} \text{ NJ}(\hat{a}, x) = \texttt{Known}; \\
%     \text{LM}(D_{\text{ref}}, x), &  \text{ otherwise.}
%     \end{cases} 
% \end{equation*}

% For a claim $c_i$ and its corresponding generated query $q_{c_i}$, we construct the retrieved 

% During question-level retrieval necessity judgment, we first generate heuristic answers using a proxy model. Then we input the heuristic answers and user questions into the retrieval necessity judgment model. If the retrieval necessity judgment model output is \texttt{Known}, we directly infer the output using the LLM as a reader for the user question. Otherwise, we use RAG:

% The necessity for retrieval for the question does not mean every aspect of the question requires search enhancement; therefore, we use claim-level retrieval necessity judgment. As for claim-level retrieval necessity judgment, we first use the query rewriting model to obtain multiple queries, and further, apply the retrieval necessity judgment model to filter queries that require search. Here we input the query and the claim that produced this query. The set of queries is given by:

\begin{table*}
\small
\centering
\setlength{\tabcolsep}{1.3mm}{
\begin{tabular}{lccccccccccc}
\toprule
\multirow{2}{*}{\textbf{Method}}&\multirow{2}{*}{\textbf{\#API}} & \multicolumn{2}{c}{\textbf{ASQA}} & \multicolumn{2}{c}{\textbf{NQ}} & \multicolumn{2}{c}{\textbf{Trivia-QA}} & \textbf{MuSiQue} & \multicolumn{3}{c}{\textbf{ELI5}} \\
\cmidrule(lr){3-4}\cmidrule(lr){5-6}\cmidrule(lr){7-8}\cmidrule(lr){9-9}\cmidrule(lr){10-12}
&& EM & Hit@1 & EM & Hit@1 & EM & Hit@1 & EM & ROUGE-1 & ROUGE-2 & ROUGE-L \\
\midrule
\multicolumn{12}{c}{Llama2-70B-Chat without Retrieval} \\
\midrule
Vanilla Chat  & 1     & 29.68 & 62.50  & 40.49 & 55.00  & 27.44 & 90.75 & 11.50   & 28.66  & 4.88   & 14.27 \\
CoT           & 1     & 26.21 & 54.50  & 35.36 & 48.75 & 23.50  & 79.00  & 11.50   & 28.12  & 4.73   & 14.06 \\
\midrule
\multicolumn{12}{c}{{Llama2-70B-Chat with Retrieval}} \\
\midrule
Direct RAG    & 1     & 27.63 & 58.00  & 42.40  & 56.00  & 28.07 & 92.25 & 10.50   & 28.61  & 4.76   & \textbf{15.76} \\
FLARE         & 2.10   & 30.08 & 63.50  & 41.36 & 55.75 & 27.41 & 89.50  & 11.25  & 27.95  & 4.72   & 13.91 \\
Self-Eval     & 2     & 29.45 & 60.75 & 42.15 & 55.75 & 27.58 & 91.50  & 10.25  & 28.70   & 4.83   & 15.39 \\
Self-Ask      & 2.67  & 26.37 & 60.25 & 38.56 & 53.00  & 26.56 & 89.50  & 9.50    & -      & -      & - \\
ITER-RETGEN   & 3     & 30.15 & 60.50  & 42.85 & 55.50  & 28.31 & 91.00  & 13.00   & 28.44  & 4.74   & 15.72 \\
SKR-KNN       & 1     & 29.38 & 61.75 & 41.90  & 55.75 & 28.16 & \textbf{92.25} & 10.25  & 28.71  & 4.80    & 15.73 \\
% \hline
SlimPLM (Ours)      & 1     & \textbf{30.73} & \textbf{65.00}  & \textbf{47.43} & \textbf{62.25} & \textbf{28.35} & 92.00  & \textbf{13.00}   & \textbf{29.97}  & \textbf{5.61}   & 15.13 \\ 
\midrule
\midrule
\multicolumn{12}{c}{Qwen-72B-Chat without Retrieval} \\
% \multicolumn{12}{c}{Without Retrieval} \\
\midrule
Vanilla Chat & 1     & 26.65 & 58.50  & 40.38 & 53.75 & 27.82 & 90.25 & 11.75  & \textbf{30.61}  & 5.21   & 15.90 \\
CoT          & 1     & 27.74 & 59.50  & 40.49 & 53.75 & 27.62 & 91.75 & 12.75  & 29.94  & 4.94   & 14.75 \\
\midrule
\multicolumn{12}{c}{Qwen-72B-Chat with Retrieval} \\
\midrule
Direct RAG   & 1     & 25.85 & 57.00  & 41.27 & 52.75 & 26.39 & 87.75 & 7.75   & 25.93  & 4.55   & \textbf{16.74} \\
FLARE        & 2.29  & 27.68 & 59.00  & 40.89 & 54.50  & 27.10  & 88.50  & \textbf{12.75}  & 30.31  & 5.20    & 15.77 \\
Self-Eval    & 2     & 27.64 & 60.00  & 42.43 & 56.00  & 27.13 & 90.50  & 7.75   & 29.19  & 5.14   & 16.05 \\
Self-Ask     & 2.76  & 22.82 & 52.25 & 36.16 & 49.25 & 25.29 & 87.50  & 9.75   & -      & -      & - \\
ITER-RETGEN  & 3     & \textbf{29.38} & 61.50  & 43.51 & 57.50  & 27.16 & 89.75 & 12.25  & 26.15  & 4.41   & 16.52 \\
SKR-KNN      & 1     & 28.08 & 61.50  & 43.08 & 56.00  & 26.38 & 88.50 & 11.25  & 27.29  & 4.75   & 16.31 \\
SlimPLM (Ours)     & 1     & 27.97 & \textbf{62.25} & \textbf{44.07} & \textbf{57.75} & \textbf{28.03} & \textbf{92.25} & 9.75   & 29.56  & \textbf{5.91}   & 16.36 \\
\bottomrule
\end{tabular}
}
\caption{Evaluation results of SlimPLM and baselines on five QA benchmarks. \#API is the average LLM inference times. Hit@1 is the proportion of instances where at least one short answer matches. }
\label{main-results}
\end{table*}

\section{Experiments}
We conduct experiments on five widely used question-answering (QA) datasets and compare the performance of our method with several baselines.
% We also compare SlimPLM with two baselines without retrieval and six baselines across three major streams using active retrieval augmentation. 

\subsection{Datasets}
We use the following five QA datasets: 
(1) Natural Questions (NQ)~\cite{Kwiatkowski2019NaturalQA}: a dataset consisting of real user questions from Google search.
(2) Trivia-QA~\cite{Joshi2017TriviaQAAL}: a realistic text-based question answering dataset.
(3) ASQA~\cite{Stelmakh2022ASQAFQ}: a dataset targeting ambiguous questions requiring answers that integrate factual information from various sources.
(4) MuSiQue~\cite{Trivedi2021MM}: a synthetic multi-hop question-answering dataset.
(5) ELI5~\cite{Fan2019ELI5LF}: a long-form question answering dataset originated from the Reddit forum.
% (1) Natural question-answering dataset: Natural Questions (NQ)~\cite{Kwiatkowski2019NaturalQA} and Trivia-QA~\cite{Joshi2017TriviaQAAL}. (2) Ambiguous question-answering dataset:  ASQA~\cite{Stelmakh2022ASQAFQ}. (3) Multi-hop question-answering dataset: MuSiQue~\cite{Trivedi2021MM}; and the long-text question-answering dataset ELI5~\cite{Fan2019ELI5LF}. 
Due to our limited resources, we randomly sample 400 questions from the test set (if any) or validation set of each dataset as the test set for evaluation. 

\subsection{Evaluation Metrics}
For all QA tasks, LLMs can freely generate any answers. For datasets annotated with long-form answers, we employ the Rouge Score~\cite{Lin2004ROUGEAP} (ROUGE) to evaluate the quality of the generated answers by comparing them with the ground-truth ones. For datasets with short answers, we use the Exact Match (EM) metric to compare the generated answer with the golden one.
% \footnote{We refer to the evaluation scripts from \url{https://github.com/google-research/language/blob/master/language/asqa/scoring.py}} 
% of the short answer annotations within the output text of the LLM. 
If the dataset provides multiple optional short answers, we also report the proportion of instances where at least one short answer matches (Hit@1).

\subsection{Baselines}
We first select two baselines without retrieval:

(1) \textbf{Vanilla Chat}: This method directly inputs the user question into LLMs to get the answer. 

(2) \textbf{CoT Prompting}~\cite{Wei2022ChainOT}: This method introduces a prompt method that lets LLMs think step-by-step to derive the final answer.

We also consider several retrieval-augmented generation methods.
% These methods input the question with the retrieved results to LLMs and get the answer. 
They differ in time and approach for retrieval necessity judgment and construction of retrieval queries. We include more baseline implementation details in Appendix~\ref{appendix-b}.

(1) \textbf{Direct RAG}: This approach applies retrieval-augmentation for all questions and directly utilizes the user question as the search query. 

(2) \textbf{FLARE}~\cite{Jiang2023ActiveRA}: This method examines the content of each sentence generated by the LLM, and uses retrieval if the generation logits are below a threshold. FLARE uses the masked sentence as a query, wherein tokens associated with low logits are masked.

(3) \textbf{Self-Eval}~\cite{Kadavath2022LanguageM}: This method uses prompts and few-shot learning to let LLM itself decide whether it needs retrieval or not. 

(4) \textbf{Self-Ask}~\cite{Press2022MeasuringAN}: This method iteratively prompts the LLM to decide whether to generate follow-up questions as queries or generate the final answer directly.
% This method iteratively generates follow-up question, search for the generated follow-up question, and answer the generated follow-up question until the final answer can be deduced.
% \textbf{ReAct}~\cite{Yao2022ReActSR} selects one operation from reasoning, action, and observation based on the input of the current round, continuing until the final answer is obtained. \textbf{DSP}~\cite{Press2022MeasuringAN} utilizes the answer generated from the previous round's LLM with retrieval enhancement as the query for the next round, repeating this process until the final answer is achieved. 
% Regarding methods that analyze question difficulty to determine if the LLM possesses relevant knowledge, we select 

(5) \textbf{SKR-KNN}~\cite{Wang2023SelfKnowledgeGR}. It uses a dense retriever to retrieve top-$k$ nearest neighbor questions from the training set. The necessity of retrieval is determined by the number of neighboring questions that require or do not require retrieval.

% \textbf{GPT-4 R\&J} uses few-shot learning\cite{Brown2020LanguageMA} and manually crafted prompts to instruct GPT-4\cite{Achiam2023GPT4TR} to perform query rewriting and retrieval necessity judgment. 
% We proceed in two steps. In the first step, we present the user question and an example to GPT-4, which then determines whether the question requires a search and the query rewrite derived from the analysis of the question. In the second step, we provide heuristic answers from a smaller language model. GPT-4 then deconstructs these heuristic answers into factual claims and, in combination with the question and each claim, determines whether a search is needed and achieves query rewriting. The prompts we use is demonstrated in Appendix.

\subsection{Implementation Details}
We conduct experiments on two open-source LLMs, Llama2-70B-Chat~\cite{Touvron2023Llama2O} and Qwen-72b-Chat~\cite{Bai2023QwenTR}. 
% Four open-source LLMs are used as proxy models, and their sizes ranges from 1B to 7B, including Llama2-7B-Chat~\cite{Touvron2023Llama2O}, Qwen-7B-Chat~\cite{Bai2023QwenTR}, TinyLlama-1.1B-Chat~\cite{Zhang2024TinyLlamaAO}, and Phi-2~\cite{Li2023TextbooksAA}. 
The default proxy model, fine-tuned query rewriting model, and retrieval necessity judgment model are built on Llama2-7B-Chat. We build a search engine on the KILT dataset's document library, which is based on the 2019 Wikipedia mirror~\cite{Petroni2020KILTAB}. BM25~\cite{Robertson2009ThePR} is used as the retriever and $\text{E5}_{\text{base}}$~\cite{Wang2022TextEB} is employed as the reranker. More implementation details are provided in Appendix~\ref{sec:appendix}.
% For the retriever, we use BM25~\cite{Robertson2009ThePR}, and for the reranker, we employed e5-base-v2~\cite{Wang2022TextEB}. We include more implementation details in Appendix~\ref{sec:appendix}.
% In our choice of search engines, we used both a specialized search engine based on the wiki document library and the general domain search engine Bing. 
% Our self-built wiki search engine can rival the performance of the commercial search engine Bing on our selected datasets. Therefore, we did not use a stronger retriever, such as using E5 directly as the retriever.

% \noindent\textbf{Query Rewrite \& Judgment model} \space We fine-tune smaller generative language models to obtain the query rewriting model and the model for determining the necessity of retrieval. For the chat model used in unrelated reasoning, we uniformly choose Llama2-7B-Chat as the base model. We designed the format of input prompts and outputs during fine-tuning and inference processes, which are detailed in the following frame.

\subsection{Experimental Results} 
The evaluation results are shown in Table~\ref{main-results}, where we uniformly chose Llama2-7B-Chat as the proxy model, a fine-tuned query rewriting model, and a fine-tuned retrieval necessity judgment model. Generally, our SlimPLM achieves superior or competitive performance on all datasets. This clearly demonstrates the effectiveness of our method. Besides, we have the following observations:

% For users' questions, our method can effectively determine whether the model needs to perform a retrieval the question and, if retrieval is necessary, what knowledge the model is missing that should be searched for.

% \subsection{Overall Comparison with Baselines} \space 

(1) On most datasets, retrieval-augmented generation methods can outperform the methods without using retrieval. This clearly demonstrates the benefit of incorporating external knowledge into open-domain QA tasks. 

(2) Compared to methods that initiate retrieval based on the results or logits generated by LLMs (\ie, Self-Eval, Self-Ask, and FALRE), our method yields better results. This validates the superiority of our method, which employs a proxy model to determine when and what the LLM needs to retrieve. Notably, our method requires the LLM to infer only once, significantly reducing the cost of inference.

(3) Comparing methods that judge retrieval necessity merely based on user questions (SKR-KNN), our method also has advantages. By using heuristic answers, it can more accurately assess the LLM's knowledge capability and formulate queries that are more precisely tailored to the question, thereby improving overall performance.

\begin{figure}
    \centering
    \includegraphics[width=\linewidth]{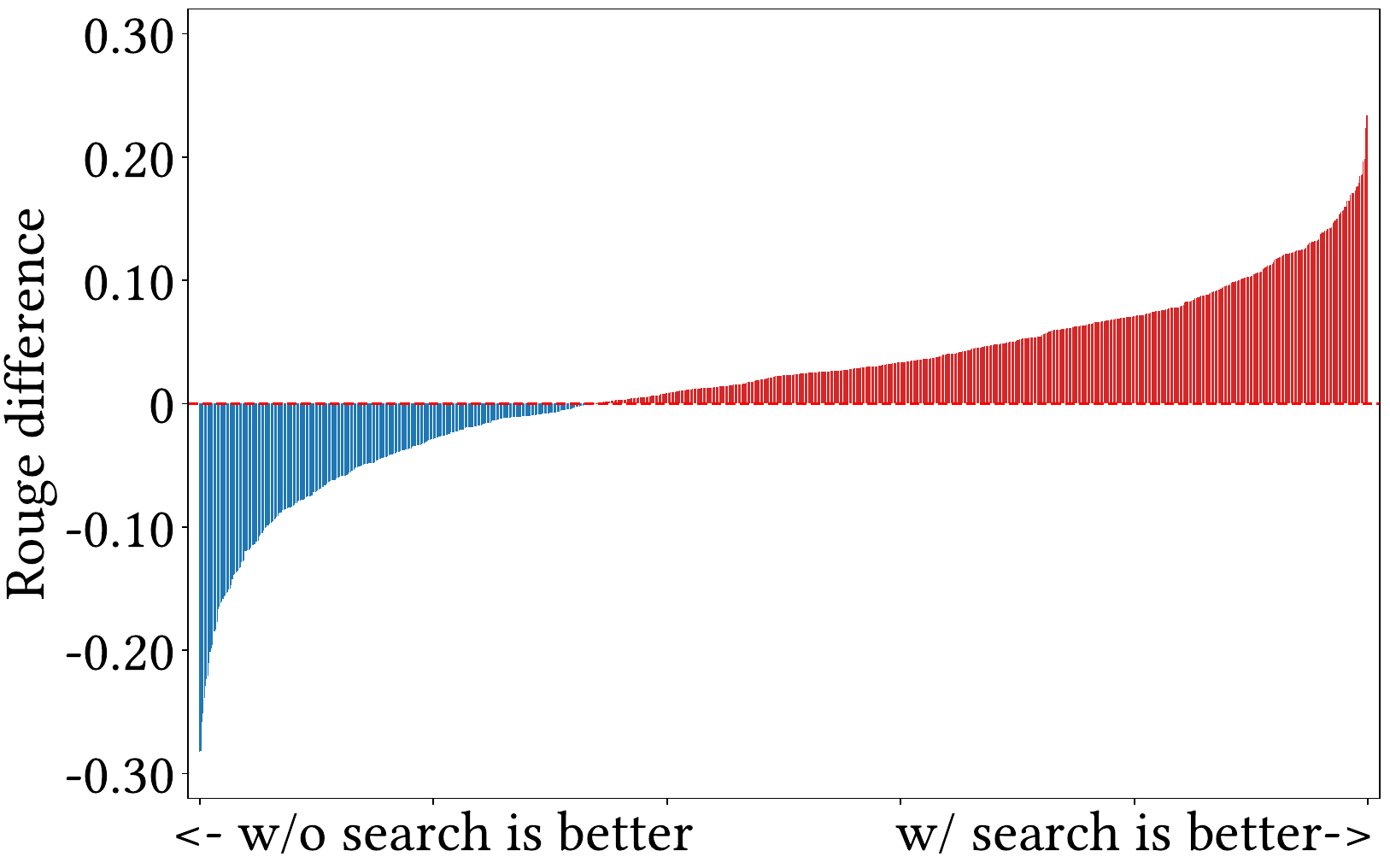}
    \caption{The rouge performance for with and without retrieval on ELI5 dataset.}
    \label{vanilla-search-diff}
\end{figure}

(4) Intriguingly, we notice that retrieval does not uniformly benefit all user questions. For example, in the ELI5 dataset, approximately 66.4\% of samples show improvement with retrieval, as shown in Figure~\ref{vanilla-search-diff}. This observation highlights the critical need to judge the necessity of retrieval. More cases where retrieval has negative impact are shown in Appendix~\ref{appendix-c}.
% there are mainly two scenarios where searching can have adverse effects: (1) The references retrieved is misleading, the LLM is provided incorrect references; (2) The references retrieved is incomplete, causing the language model to focus on the answer found and overlook other possible answers.
% \noindent\textbf{Mid-results of retrieval necessity Judge} \space 

% \noindent\textbf{Mid-results of Rewrite Model} \space 

\begin{table*}
\small
\centering
\setlength{\tabcolsep}{1.2mm}{
\begin{tabular}{lccccccc}
\toprule
\multirow{2}{*}{\textbf{Method}} & \multicolumn{2}{c}{\textbf{ASQA}} & \multicolumn{2}{c}{\textbf{NQ}} & \multicolumn{2}{c}{\textbf{Trivia-QA}} & \textbf{MuSiQue}  \\
\cmidrule(lr){2-3}\cmidrule(lr){4-5}\cmidrule(lr){6-7}\cmidrule(lr){8-8}
& EM    & Hit@1 & EM  & Hit@1 & EM  & Hit@1  & EM    \\ \midrule
SlimPLM & 30.73 & 65.00 & 47.43 & 62.25 & 28.35 & 92.00 & 13.00 \\
\quad \textit{w/o} QR & 29.19 (5.0\%$\downarrow$) & 61.75 (5.0\%$\downarrow$) & 45.16 (4.8\%$\downarrow$) & 59.75 (4.0\%$\downarrow$) & 28.07 (1.0\%$\downarrow$) & 92.50 (0.5\%$\uparrow$) & 11.50 (1.2\%$\downarrow$) \\ 
\quad \textit{w/o} QJ & 29.43 (4.2\%$\downarrow$) & 61.75 (5.0\%$\downarrow$) & 43.03 (9.3\%$\downarrow$) & 57.25 (8.0\%$\downarrow$) & 27.91 (1.6\%$\downarrow$) & 90.25 (1.9\%$\downarrow$) & 12.75 (1.9\%$\downarrow$) \\
\quad \textit{w/o} QF & 30.73 (0.0\%) & 64.75 (0.4\%$\downarrow$) & 46.62 (1.7\%$\downarrow$) & 61.25 (1.6\%$\downarrow$) & 28.27 (0.3\%$\downarrow$) & 91.75 (0.3\%$\downarrow$) & 12.50 (3.9\%$\downarrow$) \\
\bottomrule
\end{tabular}
}
\caption{Ablation study on Llama2-70B-Chat. ``QR'', ``QJ'', and ``QF'' denote the query rewriting, question-level retrieval necessity judgment, and claim-based query filter, respectively.}
% stands for  for the user question. CJ stands for claim-level retrieval necessity judgment. }
\label{query-rewrite-ablation}
\end{table*}

\begin{table*}
\small
\centering
\setlength{\tabcolsep}{1.4mm}{
\begin{tabular}{lcccccccccc}
\toprule
\multirow{2}{*}{\textbf{Method}} & \multicolumn{2}{c}{\textbf{ASQA}} & \multicolumn{2}{c}{\textbf{NQ}} & \multicolumn{2}{c}{\textbf{Trivia-QA}} & \textbf{MuSiQue} & \multicolumn{3}{c}{\textbf{ELI5}} \\
\cmidrule(lr){2-3}\cmidrule(lr){4-5}\cmidrule(lr){6-7}\cmidrule(lr){8-8}\cmidrule(lr){9-11}
& EM     & Hit@1  & EM     & Hit@1  & EM     & Hit@1  & EM     & ROUGE-1 & ROUGE-2 & ROUGE-L  \\ 
\midrule
\multicolumn{11}{c}{Llama2-70B-Chat} \\
\midrule
Vanilla Chat     & 29.68 & 62.50  & 40.49 & 55.00  & 27.44 & 90.75 & 11.50   & 28.66  & 4.88   & 14.27 \\
Llama2-7B-Chat   & 30.73 & \textbf{65.00}  & \textbf{47.43} & \textbf{62.25} & 28.35 & 92.75 & 13.00   & 29.97  & 5.61   & 15.13                                  \\
Baichuan2-7B-Chat & \textbf{31.19} & 63.25 & 44.57 & 58.75 & \textbf{28.44} & \textbf{93.25} & \textbf{14.00} & 29.95 & 5.64 & 15.49 \\
Qwen-7B-Chat    & 29.62 & 60.25 & 42.53 & 56.25 & 27.93 & 92.25 & 13.00   & 29.95  & 5.57   & \textbf{16.16}                                  \\
Phi-2 (2.7B) & 28.96 & 60.50  & 43.33 & 57.50  & 27.99 & 91.50  & 13.75 & \textbf{30.34}  & \textbf{5.82} & 15.48 \\ 
TinyLlama-1.1B-Chat & 30.47 & 60.50  & 44.24 & 56.75 & 28.02 & 91.00  & 11.50   & 30.05  & 5.56   & 15.37                                  \\

\midrule
\multicolumn{11}{c}{Qwen-72B-Chat} \\
\midrule
% Method    & EM    & Hit@1 & EM    & Hit@1 & EM & EM    & Hit@1 & rouge1 & rouge2 & rougeL  \\ \midrule
Vanilla Chat     & 26.65 & 58.50  & 40.38 & 53.75 & 27.82 & 90.25 & \textbf{11.75}  & \textbf{30.61}  & 5.21   & 15.90 \\
Llama2-7B-Chat   & 27.97 & \textbf{62.25} & \textbf{44.07} & \textbf{57.75} & \textbf{28.03} & \textbf{92.25} & 9.75   & 29.56  & \textbf{5.91}   & \textbf{16.36}      \\
Baichuan2-7B-Chat & \textbf{28.11} & 62.00 & 43.46 & 57.25 & 27.65 & 91.75 & 11.00 & 28.36 & 5.69 & 16.28 \\
Qwen-7B-Chat    & 27.76 & 59.75 & 42.54 & 55.75 & 27.22 & 90.25 & 8.75   & 29.44  & 5.74   & 16.33 \\
Phi-2 (2.7B)      & 26.95 & 59.50  & 42.22 & 54.25 & 27.10 & 89.00  & 10.75  & 29.17  & 5.83   & 16.32 \\ 
TinyLlama-1.1B-Chat & 27.61 & 58.25 & 42.36 & 55.25 & 27.67 & 91.25 & 9.25   & 28.80  & 5.64   & 16.12 \\
\bottomrule
\end{tabular}
}
\caption{Performance Comparison of Various Proxy Methods to Vanilla Chat.}
\label{different-proxy-model}
\end{table*}

\subsection{Further Analysis}
We further conduct a series of experiments to investigate the impact of different settings in our method.

\paragraph{Ablation Study}
We first examine the effectiveness of different modules in our method by an ablation study. This experiment is conducted by removing the heuristic-answer-driven query rewriting (\textit{w/o} QR), question-Level retrieval necessity judgment (\textit{w/o} RJ), and Claim-based Query Filter (\textit{w/o} QF), respectively. From the results are shown in Table~\ref{query-rewrite-ablation}, we can see: 

(1) If query rewriting is removed, then retrieval necessity judgment between vanilla chat and direct RAG is applied. Performing query rewriting can both enhance the comprehensiveness and relevance of retrieved references.

% Despite the presence of hallucinations in heuristic answers, they still can indicate the search direction for queries.
(2) When retrieval necessity judgment is removed, all questions will use retrieval results for generation. LLMs will be led astray on questions that they can perform well on their own knowledge.
% The judgment on the necessity of retrieval can enhance the effect within the RAG pipeline. Whether using methods with slightly weaker or stronger search capabilities, our model's selection to perform generation without retrieval can improve the end-to-end chat performance for a subset of cases. It is evident that if we do not filter queries and blindly increase them, it may worsen the search results, ultimately affecting the end-to-end performance.

(3) If claim-based query filter is removed, then retrieval is applied to every query derived from the heuristic answer. Not filtering queries which contain contents that do not require retrieval worsens the search results. 
% Here On the necessity of retrieval can enhance the effect within the RAG pipeline.
% Table~\ref{query-rewrite-ablation} shows that if heuristic answers undergo only claim extraction and query rewriting, without assessing the necessity of retrieval, the overall effectiveness may decline. This is because a large number of queries that do not require searching can distract the LLM with the documents they retrieve. Therefore, it is necessary to filter these queries for retrieval necessity based on the claims extracted.

\begin{figure}
    \centering
    \includegraphics[width=\linewidth]{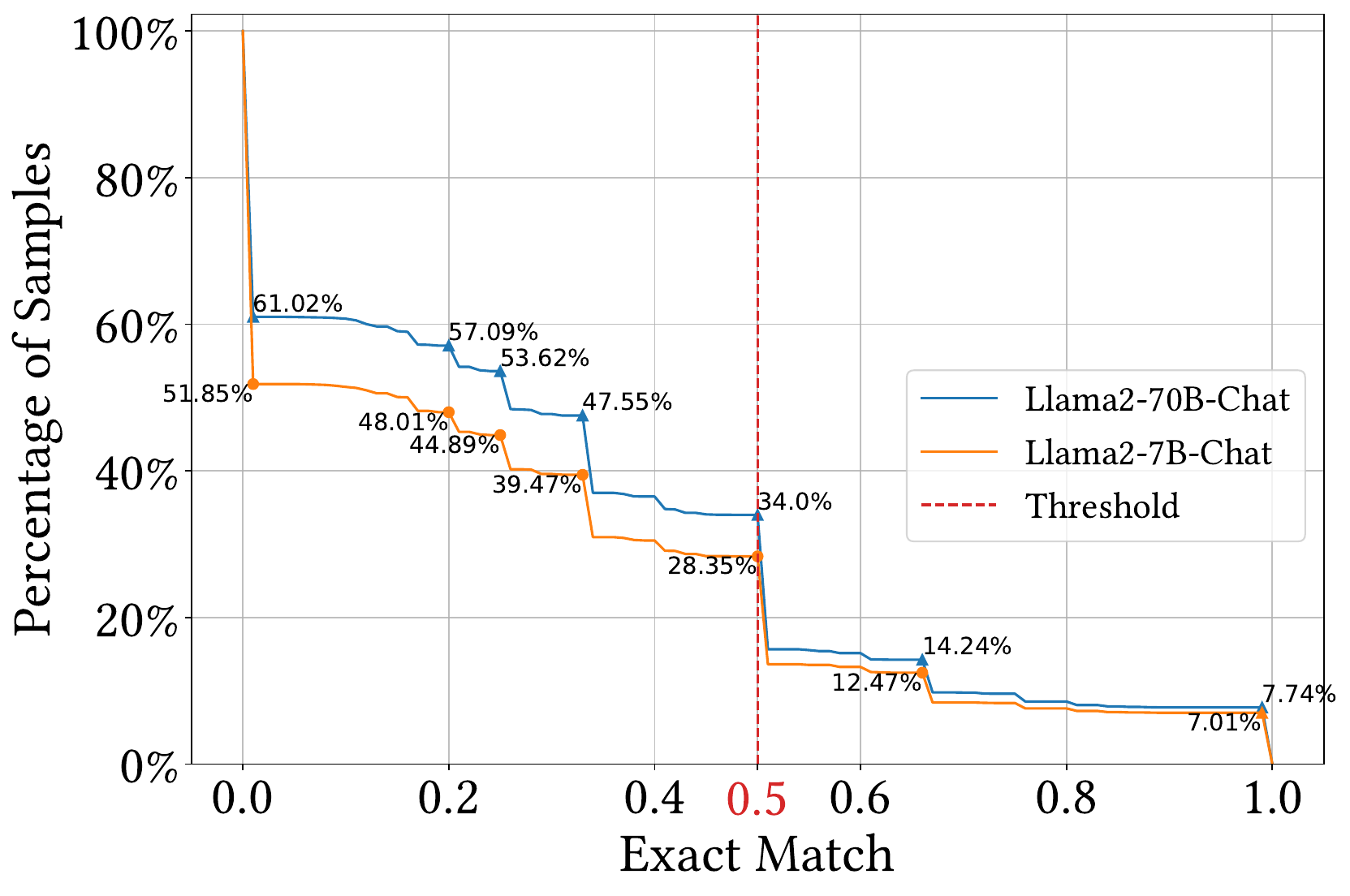}
    \caption{The proportion of samples ($y$-axis) with EM scores higher than certain values ($x$-axis).}
    % The proportion of samples where the performance of two language models surpasses the current metric size, out of the total sample population. The red dashed line represents the threshold we set for determining when a model's response is good enough without retrieval.}
    \label{slm-llm-gap}
\end{figure}

\paragraph{Knowledge Ability Consensus between Proxy Models and LLMs}
\label{knowledge-gap}
In this experiment, we compared the knowledge capabilities of LLMs and proxy models, and confirmed their consensus. Our findings can be summarized as follows: 

(1) The difference in capabilities between the proxy model and the LLM is primarily manifested in the knowledge of lower mastery levels. As illustrated in Figure~\ref{slm-llm-gap}, on the ASQA dataset, the difference between the 70B and 7B language models is very slight for samples with an EM score greater than 0.5. Their differences are primarily evident in samples with an EM score less than 0.5.

(2) The higher the level of some knowledge mastered by the proxy model, the higher the level of mastery by the LLM. Further experiments on ASQA shows over 82.19\% of the samples with an EM score greater than 0.5 for the 7B model overlaps with those of the 70B model.

The experimental results above offer a theoretical basis for our method. If the proxy model can correctly answer the question, then the LLM is very likely to answer it correctly as well. Applying vanilla chat for them can better leverage the inherent knowledge capabilities of LLMs.

\paragraph{Impact of Various Proxy Models}
We also explore the impact of using different proxy models in our method. This experiment is conducted by using four open-source LLMs with different sizes as the proxy model, including Llama2-7B-Chat~\cite{Touvron2023Llama2O}, Baichuan2-7B-Chat~\cite{Yang2023Baichuan2O}, Qwen-7B-Chat~\cite{Bai2023QwenTR}, and Phi-2~\cite{Li2023TextbooksAA}, TinyLlama-1.1B-Chat~\cite{Zhang2024TinyLlamaAO}. Experimental results are shown in Table~\ref{different-proxy-model}. We can see that in most datasets, Llama2-7B-Chat can provide the best results. Furthermore, Llama2-7B-Chat contributes a greater improvement to Llama2-70B-Chat than to Qwen-72B-Chat, we attribute this to the better knowledge alignment between Llama models.  
% the improvement in end-to-end question-answering performance for LLMs using our method depends on two factors: (1) the absolute inherent knowledge capability of the proxy model; (2) the similarity in knowledge capability between the LLM and the proxy model.

\begin{table}
\small
\centering
\setlength{\tabcolsep}{1.7mm}{
\begin{tabular}{lccccc}
\toprule
\textbf{Dataset}    & \textbf{Chat}   & \textbf{Proxy} & \textbf{Rewrite} & \textbf{Judge} & \textbf{Total}\\
\midrule
ASQA       & 192.86 & 24.42     & 35.27   & 3.38  & 63.07 \\ 
NQ         & 249.74 & 29.61     & 38.65   & 3.65  & 71.91 \\ 
TQA        & 114.07 & 15.73     & 28.13   & 2.47  & 46.33 \\ 
MuSiQ    & 168.47 & 19.00      & 30.84   & 2.36  & 52.20  \\ 
ELI5       & 471.22 & 47.82     & 46.82   & 4.23  & 98.87 \\ 
\bottomrule
\end{tabular}
}
\caption{The number of tokens used by LLM, and the additional tokens brought by components of SlimPLM separately and in total(Total). Components includes proxy model (Proxy), query rewriting model((Rewrite), and search necessity judge model(Judge). }
\label{token}
\end{table}

\paragraph{Computational Cost Analysis}
In our method, we use a proxy model, a query rewriting model, and a retrieval necessity judgment model based on relatively smaller LLMs (Llama2-7B-Chat). To investigate their computational efficiency, we analyze the average number of tokens generated by each model and calculate the associated costs.  This calculation is based on the assumption that the computational expense per token for a 7B model is roughly 1/10 that of a 70B model—--a conservative estimate, given that the actual cost differential is likely to exceed this ratio~\cite{Kaplan2020ScalingLF}. Table~\ref{token} lists the additional computational costs required by each component and the total cost. The analysis reveals that the additional costs are substantially lower (1/4 to 1/3) compared to the costs of a single inference by an LLM. This observation validates the economic advantages of our method.
% The results indicate that the additional costs of our method are significantly lower (about 1/4 to 1/3) than those of a single time LLM inference. 
% As shown in Table~\ref{token}, we have calculated the additional computational cost required by our method. Our proxy model (Proxy), query rewriting model (Rewrite), and retrieval necessity judgment model (Judge) each have a parameter size of 7B. The per-token computational cost for 7B is approximately 1/10 that of a 70B chat model(Chat). 
% Based on this standard, we have estimated the additional computational cost required for each of our modules and the total additional cost. Experimental evidence suggests that our method essentially requires only about 1/4 to 1/3 of the additional computational cost compared to vanilla chat.

\section{Conclusion}
In conclusion, our research proposes a new paradigm for RAG, utilizing a smaller LLM as proxy model. Based on the heuristic answer by proxy model, we conduct query rewriting, retrieval necessity judgment, and claim-based query filtering. This approach enables accurate perception for when and what to retrieve for LLMs. Experiments across various datasets show a marked improvement in the end-to-end performance of LLM question-answering, achieving or exceeding state-of-the-art results. Moreover, this enhancement is attained with little additional computational cost.

\section*{Limitations}
In scenarios where almost all user questions are primarily outside the scope of the LLM's pre-training corpus, or where almost all the questions do not require external knowledge, our method proves challenging to utilize. In these situations, opting either for a full retrieval or without retrieval at all may be a more suitable approach. Additionally, we acknowledge a gap in the knowledge capabilities between proxy models and LLMs. Heuristic answers are unable to fully reflect the true knowledge capability of the LLMs. Moreover, our current method employs three models: a proxy model, a query rewriting model, and a retrieval necessity judgment model. The pipeline appears somewhat complex; integrating these functions into a single generative framework would be preferable.

\section*{Acknowledgments}
This work was supported by Beijing Natural Science Foundation No. L233008, CCF-BaiChuan-Ebtech Foundation Model Fund, National Natural Science Foundation of China No. 62272467,  the fund for building world-class universities (disciplines) of Renmin University of China, and Public Computing Cloud, Renmin University of China. The work was partially done at the Engineering Research Center of Next-Generation Intelligent Search and Recommendation, MOE.

\newpage
\appendix

\section{SlimPLM Implementation Details}
\label{sec:appendix}

\noindent\textbf{Model Fine-tuning} \space Our query rewriting model and the retrieval necessity judgment model are both obtained by instruction fine-tuning from Llama2-7B-Chat. We find that models fine-tuned with data collected from datasets annotated with multiple short answers possess better generalization abilities. They can adapt to various tasks including ambiguous QA, natural questions, long-form QA, and rewritten queries. We collect 5000 samples each from the training sets of ASQA, Natural Questions, and Trivia-QA. Through rule-based filtering, we formed the fine-tuning data for the retrieval necessity judgment model, as shown in Table~\ref{necessity-judge-sample}. Because the number of unknown samples significantly exceeds that of known samples, we downsample the unknown samples to make their proportions roughly equal. 
For the query rewriting model, we collecte 1000 samples each from ASQA, Natural Questions, Trivia-QA, MuSiQue, ELI5, and then use GPT-4 for auxiliary annotation. The prompt we use to induce GPT-4 annotation is displayed in Table~\ref{query-rewrite-prompt}.

\noindent \textbf{RAG Prompts} \space 
RAG prompts concatenate the reference document in front of the question for enhanced retrieval generation. For datasets annotated with short-form answers and long-form answers, we use different RAG prompts. This is because short-form QA requires the completeness of answers, while long-answer QA demands the fluency of answers. Prompts we use are demonstrated in Table~\ref{short-long-prompt}. We apply the same prompt strategy across all baselines unless some methods have very strict requirements for prompts, such as Self-Ask~\cite{Press2022MeasuringAN} and FLARE~\cite{Jiang2023ActiveRA}.

\section{Baseline Implementation Details}
\label{appendix-b}
For methods that require multiple rounds of large language model reasoning, we observe that three rounds of reasoning can already solve most of the problems in our dataset. Methods with an indefinite number of reasoning rounds (Self-Ask~\cite{Press2022MeasuringAN}, FLARE~\cite{Jiang2023ActiveRA}) mostly stop iterating after three rounds. Considering the limitations of computational resources, we set the maximum number of iterations to three rounds. We also set the iteration count of 3 for ITER-RETGEN~\cite{Shao2023EnhancingRL}.

The results of Self-Ask~\cite{Press2022MeasuringAN} on the ELI5 dataset are not compared is that Self-Ask can only output short answers due to prompt limitations, which does not meet the ELI5 setting for long text annotations.

The special prompt for FLARE is demonstrated in Table~\ref{flare-prompt}.

The special prompt for Self-Ask is demonstrated in Table~\ref{selfask-prompt}. Specifically, we use the LLM itself as a reader to extract concise answers as intermediate answers from the documents found in search. This was implemented using the Google API in the original paper, but we use our own Wiki document search library, hence the need for this approach.

\section{Case Study}
\label{appendix-c}
We provide some cases of misleading references in Table~\ref{references}. There are mainly two scenarios where searching can have adverse effects: (1) The references retrieved is misleading, the LLM is provided incorrect references; (2) The references retrieved is incomplete, causing the language model to focus on the answer found and overlook other possible answers.

\begin{table}
\small
\centering
\begin{tabular}{lccc}
\toprule
\textbf{Dataset}    & \textbf{Known}   & \textbf{Unknown} & \textbf{Dropped} \\ 
\midrule
ASQA       & 593 & 592 & 3,168 \\ 
NQ         & 1,873 & 1,874 & 1,253 \\ 
TQA        & 588 & 588 & 3,824 \\ 
\bottomrule
\end{tabular}
\caption{The number of unknown, known and dropped samples for retrieval necessity judgment model.}
\label{necessity-judge-sample}
\end{table}

\begin{table}
\centering
\small
\begin{tabular}{l}
\toprule
\textbf{RAG Prompt for FLARE} \\
\midrule
Search results: \\
{}[1]\textbf{ doc 1} \\
{}[2]\textbf{ doc 2} \\
... \\
\textbf{question} \\
\bottomrule
\end{tabular}
\caption{RAG prompt for FLARE.}
\label{flare-prompt}
\end{table}

\begin{table*}
\centering
\small
\begin{tabular}{L{16cm}}
\toprule
\textbf{Question:}
Where are the winter Olympics and when do they start? \\
\midrule

\textbf{reference:}
Åre and Östersund, Sweden will host the next World Winter Games between February 2 to 13, 2021. It will mark the first time that Sweden has ever hosted the Special Olympics World Games.

\textbf{Error Type:} 
irrelevant reference 

\textbf{reference:}
The EOC launched the bid process on 20 September 2018 after a meeting of the constituent National Olympic Committees in Stockholm.

\textbf{Error Type:} 
irrelevant reference \\
\midrule
\textbf{Question:}
When did the golden state warriors win the finals? \\
\midrule

\textbf{reference:}
The 2017 NBA playoffs began on April 15, 2017. It concluded with the Golden State Warriors defeating the Cleveland Cavaliers 4 games to 1 in the NBA Finals, their third consecutive meeting at the Finals.

\textbf{Error Type:} 
incomplete reference 

\textbf{reference:}
This Finals was the first time in NBA history the same two teams had met for a third consecutive year. The Cavaliers sought to repeat as champions after winning the championship in 2016, while the Warriors won the first meeting in 2015.

\textbf{Error Type:} 
incomplete reference \\

\bottomrule
\end{tabular}
\caption{Cases of misleading and incomplete references.}
\label{references}
\end{table*}

\begin{table*}
\centering
\small
\begin{tabular}{L{16cm}}
\toprule
\textbf{GPT-4 Prompt for Annotating Query Rewrite from User Question} \\

\midrule
Your task is to perform text analysis on user conversations, and complete the last json item. You need to follow the following rules:

1. Classify user conversations into the following categories: text rewriting, mathematical problems, knowledge questions, text creation, table processing, translation, summarization, logical reasoning, open qa, coding, text classification, information extraction, brainstorming, exams, role-playing, others. The format should be a string and stored in the task field.

2. Determine whether the answer of user input is closely related to current datetime, and store it in the timeliness field in boolean format.

3. If the user’s request involves reasoning, each reasoning process should be described as questions and split into as many sub-questions as possible.

4. The sub-questions after splitting should be placed in the question field in questions, and the sub-questions should be fully described without using pronouns such as “he”, “this”, or “that”.

5. If the sub-question involves very strict factual information such as personal relationships, time, location, policies, regulations, etc., which requires the use of a search engine to answer, then it needs to be marked as needSearch=true, and the generated search term should be placed in searchWord.

6. If the sub-question is a chit-chat question such as "how are you" or a pure mathematical problem, coding, logical reasoning, creative thinking, or common sense problem, then no search is needed.

7. Extract the entities and events involved in the user's request and store them in the entities and events fields respectively. The format is a list of strings. Note that the entities and events should be higly informative, and should not be a user instruction or a question. \\
\midrule
\textbf{GPT-4 Prompt for Annotating Query Rewrite from User Question} \\
\midrule
<<SYS>>You are asked to first separate a given text by claims and then provide a search query to verify each claim if needed.
Here are some requirements:
1. The separation is conducted according to the meaning and each claim should be be brief and contain as one key claim.
2. Do not add any hallucinated information or miss any information.
3. The claims should be independent and self-contained, and the claims should be fully described without using pronouns such as “he”, “this”, or “that”.
4. The query is derived from it's corresponding claim and the original user question, and should be useful to check the factuality of the claim.
5. If the claim does not contain any fact relevant with the original user question, or only contains simple commen senses, then search is not required.
6. The final return should strictly follow the given format.
Like this: <Claims> <Claim(claim1)> <Search(True/False)> <Query(query1)> <Claim(claim2)> <Search(True/False)> <Query(query2)> <Claim(claim3)><Search(True/False)><Query(query3)>......</Claims> <</SYS>> \\
\bottomrule
\end{tabular}
\caption{The prompt to induce GPT-4 auxiliary annotation for query rewriting model.}
\label{query-rewrite-prompt}
\end{table*}

\begin{table*}
\centering
\small
\begin{tabular}{L{16cm}}
\toprule
\textbf{RAG Prompt for Short-Form QA} \\

\midrule
<<SYS>>

Now, based on the following reference and your knowledge, please answer the question more succinctly and professionally. The reference is delimited by triple brackets [[[]]]. The question is delimited by triple parentheses ((())). You should include as many possible answers as you can. 

<</SYS>>

Reference: [[[{\textbf{reference}}]]], 

question: ((({\textbf{question}}))) \\
\midrule
\textbf{RAG Prompt for Long-form QA} \\
\midrule
<<SYS>>

Now, based on the following reference and your knowledge, please answer the question more succinctly and professionally. The reference is delimited by triple brackets [[[]]]. The question is delimited by triple parentheses ((())). You are not allowed to add fabrications or hallucinations.

<</SYS>>

Reference: [[[{\textbf{reference}}]]], 

question: ((({\textbf{question}}))) \\
\bottomrule
\end{tabular}
\caption{RAG prompt for different tasks.}
\label{short-long-prompt}
\end{table*}

\begin{table*}
\centering
\small
\begin{tabular}{L{16cm}}
\toprule
\textbf{RAG Prompt for Self-Ask} \\

\midrule
<<SYS>>

Given the following question, answer it by providing follow up questions and intermediate answers. If no follow up questions are necessary, answer the question directly.

<<SYS>>

Question: Who lived longer, Muhammad Ali or Alan Turing?

Are follow up questions needed here: Yes.

Follow up: How old was Muhammad Ali when he died?

Intermediate answer: Muhammad Ali was 74 years old when he died.

Follow up: How old was Alan Turing when he died?

Intermediate answer: Alan Turing was 41 years old when he died.

So the final answer is: Muhammad Ali 

Question: When was the founder of craigslist born?

Are follow up questions needed here: Yes.

Follow up: Who was the founder of craigslist?

Intermediate answer: Craigslist was founded by Craig Newmark.

Follow up: When was Craig Newmark born?

Intermediate answer: Craig Newmark was born on December 6, 1952.

So the final answer is: December 6, 1952

Question: \textbf{question}\\
\midrule
\textbf{RAG Prompt for Self-Ask Reference Reader} \\
\midrule
Given the following reference, answer it by a brief sentence. You are not allowed to add fabrications or hallucinations.

\textbf{reference}

Question: How old was Muhammad Ali when he died?

Answer: Muhammad Ali was 74 years old when he died.

Question: Who was the founder of craigslist?

Answer: Craigslist was founded by Craig Newmark.

Question: Who was the father of Mary Ball Washington?

Answer: The father of Mary Ball Washington was Joseph Ball.

Question: Who is the director of Casino Royale? 

Answer: The director of Casino Royale is Martin Campbell. 

Question: \textbf{question}

Answer: \\
\bottomrule
\end{tabular}
\caption{RAG prompt for Self-Ask.}
\label{selfask-prompt}
\end{table*}


\begin{thebibliography}{75}
\expandafter\ifx\csname natexlab\endcsname\relax\def\natexlab#1{#1}\fi

\bibitem[{Almazrouei et~al.(2023)Almazrouei, Alobeidli, Alshamsi, Cappelli, Cojocaru, Debbah, Goffinet, Hesslow, Launay, Malartic, Mazzotta, Noune, Pannier, and Penedo}]{Almazrouei2023TheFS}
Ebtesam Almazrouei, Hamza Alobeidli, Abdulaziz Alshamsi, Alessandro Cappelli, Ruxandra Cojocaru, M{\'{e}}rouane Debbah, {\'{E}}tienne Goffinet, Daniel Hesslow, Julien Launay, Quentin Malartic, Daniele Mazzotta, Badreddine Noune, Baptiste Pannier, and Guilherme Penedo. 2023.
\newblock \href {https://doi.org/10.48550/ARXIV.2311.16867} {The falcon series of open language models}.
\newblock \emph{CoRR}, abs/2311.16867.

\bibitem[{Asai et~al.(2023)Asai, Wu, Wang, Sil, and Hajishirzi}]{Asai2023SelfRAGLT}
Akari Asai, Zeqiu Wu, Yizhong Wang, Avirup Sil, and Hannaneh Hajishirzi. 2023.
\newblock \href {https://doi.org/10.48550/ARXIV.2310.11511} {Self-rag: Learning to retrieve, generate, and critique through self-reflection}.
\newblock \emph{CoRR}, abs/2310.11511.

\bibitem[{Bai et~al.(2023)Bai, Bai, Chu, Cui, Dang, Deng, Fan, Ge, Han, Huang, Hui, Ji, Li, Lin, Lin, Liu, Liu, Lu, Lu, Ma, Men, Ren, Ren, Tan, Tan, Tu, Wang, Wang, Wang, Wu, Xu, Xu, Yang, Yang, Yang, Yang, Yao, Yu, Yuan, Yuan, Zhang, Zhang, Zhang, Zhang, Zhou, Zhou, Zhou, and Zhu}]{Bai2023QwenTR}
Jinze Bai, Shuai Bai, Yunfei Chu, Zeyu Cui, Kai Dang, Xiaodong Deng, Yang Fan, Wenbin Ge, Yu~Han, Fei Huang, Binyuan Hui, Luo Ji, Mei Li, Junyang Lin, Runji Lin, Dayiheng Liu, Gao Liu, Chengqiang Lu, Keming Lu, Jianxin Ma, Rui Men, Xingzhang Ren, Xuancheng Ren, Chuanqi Tan, Sinan Tan, Jianhong Tu, Peng Wang, Shijie Wang, Wei Wang, Shengguang Wu, Benfeng Xu, Jin Xu, An~Yang, Hao Yang, Jian Yang, Shusheng Yang, Yang Yao, Bowen Yu, Hongyi Yuan, Zheng Yuan, Jianwei Zhang, Xingxuan Zhang, Yichang Zhang, Zhenru Zhang, Chang Zhou, Jingren Zhou, Xiaohuan Zhou, and Tianhang Zhu. 2023.
\newblock \href {https://doi.org/10.48550/ARXIV.2309.16609} {Qwen technical report}.
\newblock \emph{CoRR}, abs/2309.16609.

\bibitem[{Chung et~al.(2022)Chung, Hou, Longpre, Zoph, Tay, Fedus, Li, Wang, Dehghani, Brahma, Webson, Gu, Dai, Suzgun, Chen, Chowdhery, Narang, Mishra, Yu, Zhao, Huang, Dai, Yu, Petrov, Chi, Dean, Devlin, Roberts, Zhou, Le, and Wei}]{Chung2022ScalingIL}
Hyung~Won Chung, Le~Hou, Shayne Longpre, Barret Zoph, Yi~Tay, William Fedus, Eric Li, Xuezhi Wang, Mostafa Dehghani, Siddhartha Brahma, Albert Webson, Shixiang~Shane Gu, Zhuyun Dai, Mirac Suzgun, Xinyun Chen, Aakanksha Chowdhery, Sharan Narang, Gaurav Mishra, Adams Yu, Vincent~Y. Zhao, Yanping Huang, Andrew~M. Dai, Hongkun Yu, Slav Petrov, Ed~H. Chi, Jeff Dean, Jacob Devlin, Adam Roberts, Denny Zhou, Quoc~V. Le, and Jason Wei. 2022.
\newblock \href {https://doi.org/10.48550/ARXIV.2210.11416} {Scaling instruction-finetuned language models}.
\newblock \emph{CoRR}, abs/2210.11416.

\bibitem[{Dhuliawala et~al.(2023)Dhuliawala, Komeili, Xu, Raileanu, Li, Celikyilmaz, and Weston}]{Dhuliawala2023ChainofVerificationRH}
Shehzaad Dhuliawala, Mojtaba Komeili, Jing Xu, Roberta Raileanu, Xian Li, Asli Celikyilmaz, and Jason Weston. 2023.
\newblock \href {https://doi.org/10.48550/ARXIV.2309.11495} {Chain-of-verification reduces hallucination in large language models}.
\newblock \emph{CoRR}, abs/2309.11495.

\bibitem[{Fan et~al.(2019)Fan, Jernite, Perez, Grangier, Weston, and Auli}]{Fan2019ELI5LF}
Angela Fan, Yacine Jernite, Ethan Perez, David Grangier, Jason Weston, and Michael Auli. 2019.
\newblock \href {https://doi.org/10.18653/V1/P19-1346} {{ELI5:} long form question answering}.
\newblock In \emph{Proceedings of the 57th Conference of the Association for Computational Linguistics, {ACL} 2019, Florence, Italy, July 28- August 2, 2019, Volume 1: Long Papers}, pages 3558--3567. Association for Computational Linguistics.

\bibitem[{Gao et~al.(2021)Gao, Biderman, Black, Golding, Hoppe, Foster, Phang, He, Thite, Nabeshima, Presser, and Leahy}]{Gao2020ThePA}
Leo Gao, Stella Biderman, Sid Black, Laurence Golding, Travis Hoppe, Charles Foster, Jason Phang, Horace He, Anish Thite, Noa Nabeshima, Shawn Presser, and Connor Leahy. 2021.
\newblock \href {http://arxiv.org/abs/2101.00027} {The pile: An 800gb dataset of diverse text for language modeling}.
\newblock \emph{CoRR}, abs/2101.00027.

\bibitem[{Gao et~al.(2023)Gao, Ma, Lin, and Callan}]{Gao2022PreciseZD}
Luyu Gao, Xueguang Ma, Jimmy Lin, and Jamie Callan. 2023.
\newblock \href {https://doi.org/10.18653/V1/2023.ACL-LONG.99} {Precise zero-shot dense retrieval without relevance labels}.
\newblock In \emph{Proceedings of the 61st Annual Meeting of the Association for Computational Linguistics (Volume 1: Long Papers), {ACL} 2023, Toronto, Canada, July 9-14, 2023}, pages 1762--1777. Association for Computational Linguistics.

\bibitem[{Guo et~al.(2017)Guo, Pleiss, Sun, and Weinberger}]{Guo2017OnCO}
Chuan Guo, Geoff Pleiss, Yu~Sun, and Kilian~Q. Weinberger. 2017.
\newblock \href {http://proceedings.mlr.press/v70/guo17a.html} {On calibration of modern neural networks}.
\newblock In \emph{Proceedings of the 34th International Conference on Machine Learning, {ICML} 2017, Sydney, NSW, Australia, 6-11 August 2017}, volume~70 of \emph{Proceedings of Machine Learning Research}, pages 1321--1330. {PMLR}.

\bibitem[{Guu et~al.(2020)Guu, Lee, Tung, Pasupat, and Chang}]{Guu2020REALMRL}
Kelvin Guu, Kenton Lee, Zora Tung, Panupong Pasupat, and Ming{-}Wei Chang. 2020.
\newblock \href {http://arxiv.org/abs/2002.08909} {{REALM:} retrieval-augmented language model pre-training}.
\newblock \emph{CoRR}, abs/2002.08909.

\bibitem[{He et~al.(2023)He, Zhang, and Roth}]{He2022RethinkingWR}
Hangfeng He, Hongming Zhang, and Dan Roth. 2023.
\newblock \href {https://doi.org/10.48550/ARXIV.2301.00303} {Rethinking with retrieval: Faithful large language model inference}.
\newblock \emph{CoRR}, abs/2301.00303.

\bibitem[{Huyen(2019)}]{chip2019evaluation}
Chip Huyen. 2019.
\newblock Evaluation metrics for language modeling.
\newblock \emph{The Gradient}.

\bibitem[{Izacard and Grave(2021)}]{DBLP:conf/eacl/IzacardG21}
Gautier Izacard and Edouard Grave. 2021.
\newblock \href {https://doi.org/10.18653/V1/2021.EACL-MAIN.74} {Leveraging passage retrieval with generative models for open domain question answering}.
\newblock In \emph{Proceedings of the 16th Conference of the European Chapter of the Association for Computational Linguistics: Main Volume, {EACL} 2021, Online, April 19 - 23, 2021}, pages 874--880. Association for Computational Linguistics.

\bibitem[{Jiang et~al.(2021)Jiang, Araki, Ding, and Neubig}]{Jiang2020HowCW}
Zhengbao Jiang, Jun Araki, Haibo Ding, and Graham Neubig. 2021.
\newblock \href {https://doi.org/10.1162/TACL\_A\_00407} {How can we know \emph{When} language models know? on the calibration of language models for question answering}.
\newblock \emph{Trans. Assoc. Comput. Linguistics}, 9:962--977.

\bibitem[{Jiang et~al.(2023)Jiang, Xu, Gao, Sun, Liu, Dwivedi{-}Yu, Yang, Callan, and Neubig}]{Jiang2023ActiveRA}
Zhengbao Jiang, Frank~F. Xu, Luyu Gao, Zhiqing Sun, Qian Liu, Jane Dwivedi{-}Yu, Yiming Yang, Jamie Callan, and Graham Neubig. 2023.
\newblock \href {https://doi.org/10.18653/V1/2023.EMNLP-MAIN.495} {Active retrieval augmented generation}.
\newblock In \emph{Proceedings of the 2023 Conference on Empirical Methods in Natural Language Processing, {EMNLP} 2023, Singapore, December 6-10, 2023}, pages 7969--7992. Association for Computational Linguistics.

\bibitem[{Jin et~al.(2024)Jin, Zhu, Zhou, and Dou}]{bider}
Jiajie Jin, Yutao Zhu, Yujia Zhou, and Zhicheng Dou. 2024.
\newblock \href {https://doi.org/10.48550/ARXIV.2402.12174} {{BIDER:} bridging knowledge inconsistency for efficient retrieval-augmented llms via key supporting evidence}.
\newblock \emph{CoRR}, abs/2402.12174.

\bibitem[{Joshi et~al.(2017)Joshi, Choi, Weld, and Zettlemoyer}]{Joshi2017TriviaQAAL}
Mandar Joshi, Eunsol Choi, Daniel~S. Weld, and Luke Zettlemoyer. 2017.
\newblock \href {https://doi.org/10.18653/V1/P17-1147} {Triviaqa: {A} large scale distantly supervised challenge dataset for reading comprehension}.
\newblock In \emph{Proceedings of the 55th Annual Meeting of the Association for Computational Linguistics, {ACL} 2017, Vancouver, Canada, July 30 - August 4, Volume 1: Long Papers}, pages 1601--1611. Association for Computational Linguistics.

\bibitem[{Kadavath et~al.(2022)Kadavath, Conerly, Askell, Henighan, Drain, Perez, Schiefer, Hatfield{-}Dodds, DasSarma, Tran{-}Johnson, Johnston, Showk, Jones, Elhage, Hume, Chen, Bai, Bowman, Fort, Ganguli, Hernandez, Jacobson, Kernion, Kravec, Lovitt, Ndousse, Olsson, Ringer, Amodei, Brown, Clark, Joseph, Mann, McCandlish, Olah, and Kaplan}]{Kadavath2022LanguageM}
Saurav Kadavath, Tom Conerly, Amanda Askell, Tom Henighan, Dawn Drain, Ethan Perez, Nicholas Schiefer, Zac Hatfield{-}Dodds, Nova DasSarma, Eli Tran{-}Johnson, Scott Johnston, Sheer~El Showk, Andy Jones, Nelson Elhage, Tristan Hume, Anna Chen, Yuntao Bai, Sam Bowman, Stanislav Fort, Deep Ganguli, Danny Hernandez, Josh Jacobson, Jackson Kernion, Shauna Kravec, Liane Lovitt, Kamal Ndousse, Catherine Olsson, Sam Ringer, Dario Amodei, Tom Brown, Jack Clark, Nicholas Joseph, Ben Mann, Sam McCandlish, Chris Olah, and Jared Kaplan. 2022.
\newblock \href {https://doi.org/10.48550/ARXIV.2207.05221} {Language models (mostly) know what they know}.
\newblock \emph{CoRR}, abs/2207.05221.

\bibitem[{Kaplan et~al.(2020)Kaplan, McCandlish, Henighan, Brown, Chess, Child, Gray, Radford, Wu, and Amodei}]{Kaplan2020ScalingLF}
Jared Kaplan, Sam McCandlish, Tom Henighan, Tom~B. Brown, Benjamin Chess, Rewon Child, Scott Gray, Alec Radford, Jeffrey Wu, and Dario Amodei. 2020.
\newblock \href {http://arxiv.org/abs/2001.08361} {Scaling laws for neural language models}.
\newblock \emph{CoRR}, abs/2001.08361.

\bibitem[{Khattab et~al.(2022)Khattab, Santhanam, Li, Hall, Liang, Potts, and Zaharia}]{Khattab2022DemonstrateSearchPredictCR}
Omar Khattab, Keshav Santhanam, Xiang~Lisa Li, David Hall, Percy Liang, Christopher Potts, and Matei Zaharia. 2022.
\newblock \href {https://doi.org/10.48550/ARXIV.2212.14024} {Demonstrate-search-predict: Composing retrieval and language models for knowledge-intensive {NLP}}.
\newblock \emph{CoRR}, abs/2212.14024.

\bibitem[{Kotha et~al.(2023)Kotha, Springer, and Raghunathan}]{Kotha2023UnderstandingCF}
Suhas Kotha, Jacob~Mitchell Springer, and Aditi Raghunathan. 2023.
\newblock \href {https://doi.org/10.48550/ARXIV.2309.10105} {Understanding catastrophic forgetting in language models via implicit inference}.
\newblock \emph{CoRR}, abs/2309.10105.

\bibitem[{Kryscinski et~al.(2020)Kryscinski, McCann, Xiong, and Socher}]{Kryscinski2019EvaluatingTF}
Wojciech Kryscinski, Bryan McCann, Caiming Xiong, and Richard Socher. 2020.
\newblock \href {https://doi.org/10.18653/V1/2020.EMNLP-MAIN.750} {Evaluating the factual consistency of abstractive text summarization}.
\newblock In \emph{Proceedings of the 2020 Conference on Empirical Methods in Natural Language Processing, {EMNLP} 2020, Online, November 16-20, 2020}, pages 9332--9346. Association for Computational Linguistics.

\bibitem[{Kwiatkowski et~al.(2019)Kwiatkowski, Palomaki, Redfield, Collins, Parikh, Alberti, Epstein, Polosukhin, Devlin, Lee, Toutanova, Jones, Kelcey, Chang, Dai, Uszkoreit, Le, and Petrov}]{Kwiatkowski2019NaturalQA}
Tom Kwiatkowski, Jennimaria Palomaki, Olivia Redfield, Michael Collins, Ankur~P. Parikh, Chris Alberti, Danielle Epstein, Illia Polosukhin, Jacob Devlin, Kenton Lee, Kristina Toutanova, Llion Jones, Matthew Kelcey, Ming{-}Wei Chang, Andrew~M. Dai, Jakob Uszkoreit, Quoc Le, and Slav Petrov. 2019.
\newblock \href {https://doi.org/10.1162/TACL\_A\_00276} {Natural questions: a benchmark for question answering research}.
\newblock \emph{Trans. Assoc. Comput. Linguistics}, 7:452--466.

\bibitem[{Li et~al.(2023)Li, Bubeck, Eldan, Giorno, Gunasekar, and Lee}]{Li2023TextbooksAA}
Yuanzhi Li, S{\'{e}}bastien Bubeck, Ronen Eldan, Allie~Del Giorno, Suriya Gunasekar, and Yin~Tat Lee. 2023.
\newblock \href {https://doi.org/10.48550/ARXIV.2309.05463} {Textbooks are all you need {II:} phi-1.5 technical report}.
\newblock \emph{CoRR}, abs/2309.05463.

\bibitem[{Lin(2004)}]{Lin2004ROUGEAP}
Chin-Yew Lin. 2004.
\newblock \href {https://aclanthology.org/W04-1013} {{ROUGE}: A package for automatic evaluation of summaries}.
\newblock In \emph{Text Summarization Branches Out}, pages 74--81, Barcelona, Spain. Association for Computational Linguistics.

\bibitem[{Lin et~al.(2022)Lin, Hilton, and Evans}]{Lin2022TeachingMT}
Stephanie Lin, Jacob Hilton, and Owain Evans. 2022.
\newblock \href {https://openreview.net/forum?id=8s8K2UZGTZ} {Teaching models to express their uncertainty in words}.
\newblock \emph{Trans. Mach. Learn. Res.}, 2022.

\bibitem[{Liu et~al.(2022)Liu, Shen, Zhang, Dolan, Carin, and Chen}]{Liu2021WhatMG}
Jiachang Liu, Dinghan Shen, Yizhe Zhang, Bill Dolan, Lawrence Carin, and Weizhu Chen. 2022.
\newblock \href {https://doi.org/10.18653/V1/2022.DEELIO-1.10} {What makes good in-context examples for gpt-3?}
\newblock In \emph{Proceedings of Deep Learning Inside Out: The 3rd Workshop on Knowledge Extraction and Integration for Deep Learning Architectures, DeeLIO@ACL 2022, Dublin, Ireland and Online, May 27, 2022}, pages 100--114. Association for Computational Linguistics.

\bibitem[{Liu et~al.(2023{\natexlab{a}})Liu, Jin, Wang, Cheng, Dou, and Wen}]{Liu2023RETALLMAR}
Jiongnan Liu, Jiajie Jin, Zihan Wang, Jiehan Cheng, Zhicheng Dou, and Ji{-}Rong Wen. 2023{\natexlab{a}}.
\newblock \href {https://doi.org/10.48550/ARXIV.2306.05212} {{RETA-LLM:} {A} retrieval-augmented large language model toolkit}.
\newblock \emph{CoRR}, abs/2306.05212.

\bibitem[{Liu et~al.(2023{\natexlab{b}})Liu, Lai, Yu, Xu, Zeng, Du, Zhang, Dong, and Tang}]{Liu2023WebGLMTA}
Xiao Liu, Hanyu Lai, Hao Yu, Yifan Xu, Aohan Zeng, Zhengxiao Du, Peng Zhang, Yuxiao Dong, and Jie Tang. 2023{\natexlab{b}}.
\newblock \href {https://doi.org/10.1145/3580305.3599931} {Webglm: Towards an efficient web-enhanced question answering system with human preferences}.
\newblock In \emph{Proceedings of the 29th {ACM} {SIGKDD} Conference on Knowledge Discovery and Data Mining, {KDD} 2023, Long Beach, CA, USA, August 6-10, 2023}, pages 4549--4560. {ACM}.

\bibitem[{Ma et~al.(2023)Ma, Gong, He, Zhao, and Duan}]{Ma2023QueryRF}
Xinbei Ma, Yeyun Gong, Pengcheng He, Hai Zhao, and Nan Duan. 2023.
\newblock \href {https://doi.org/10.48550/ARXIV.2305.14283} {Query rewriting for retrieval-augmented large language models}.
\newblock \emph{CoRR}, abs/2305.14283.

\bibitem[{Mallen et~al.(2023)Mallen, Asai, Zhong, Das, Khashabi, and Hajishirzi}]{Mallen2022WhenNT}
Alex Mallen, Akari Asai, Victor Zhong, Rajarshi Das, Daniel Khashabi, and Hannaneh Hajishirzi. 2023.
\newblock \href {https://doi.org/10.18653/V1/2023.ACL-LONG.546} {When not to trust language models: Investigating effectiveness of parametric and non-parametric memories}.
\newblock In \emph{Proceedings of the 61st Annual Meeting of the Association for Computational Linguistics (Volume 1: Long Papers), {ACL} 2023, Toronto, Canada, July 9-14, 2023}, pages 9802--9822. Association for Computational Linguistics.

\bibitem[{Maynez et~al.(2020)Maynez, Narayan, Bohnet, and McDonald}]{Maynez2020OnFA}
Joshua Maynez, Shashi Narayan, Bernd Bohnet, and Ryan~T. McDonald. 2020.
\newblock \href {https://doi.org/10.18653/V1/2020.ACL-MAIN.173} {On faithfulness and factuality in abstractive summarization}.
\newblock In \emph{Proceedings of the 58th Annual Meeting of the Association for Computational Linguistics, {ACL} 2020, Online, July 5-10, 2020}, pages 1906--1919. Association for Computational Linguistics.

\bibitem[{Min et~al.(2023)Min, Krishna, Lyu, Lewis, Yih, Koh, Iyyer, Zettlemoyer, and Hajishirzi}]{Min2023FActScoreFA}
Sewon Min, Kalpesh Krishna, Xinxi Lyu, Mike Lewis, Wen{-}tau Yih, Pang~Wei Koh, Mohit Iyyer, Luke Zettlemoyer, and Hannaneh Hajishirzi. 2023.
\newblock \href {https://doi.org/10.18653/V1/2023.EMNLP-MAIN.741} {Factscore: Fine-grained atomic evaluation of factual precision in long form text generation}.
\newblock In \emph{Proceedings of the 2023 Conference on Empirical Methods in Natural Language Processing, {EMNLP} 2023, Singapore, December 6-10, 2023}, pages 12076--12100. Association for Computational Linguistics.

\bibitem[{Min et~al.(2022)Min, Lyu, Holtzman, Artetxe, Lewis, Hajishirzi, and Zettlemoyer}]{Min2022RethinkingTR}
Sewon Min, Xinxi Lyu, Ari Holtzman, Mikel Artetxe, Mike Lewis, Hannaneh Hajishirzi, and Luke Zettlemoyer. 2022.
\newblock \href {https://doi.org/10.18653/V1/2022.EMNLP-MAIN.759} {Rethinking the role of demonstrations: What makes in-context learning work?}
\newblock In \emph{Proceedings of the 2022 Conference on Empirical Methods in Natural Language Processing, {EMNLP} 2022, Abu Dhabi, United Arab Emirates, December 7-11, 2022}, pages 11048--11064. Association for Computational Linguistics.

\bibitem[{Nakano et~al.(2021)Nakano, Hilton, Balaji, Wu, Ouyang, Kim, Hesse, Jain, Kosaraju, Saunders, Jiang, Cobbe, Eloundou, Krueger, Button, Knight, Chess, and Schulman}]{Nakano2021WebGPTBQ}
Reiichiro Nakano, Jacob Hilton, Suchir Balaji, Jeff Wu, Long Ouyang, Christina Kim, Christopher Hesse, Shantanu Jain, Vineet Kosaraju, William Saunders, Xu~Jiang, Karl Cobbe, Tyna Eloundou, Gretchen Krueger, Kevin Button, Matthew Knight, Benjamin Chess, and John Schulman. 2021.
\newblock \href {http://arxiv.org/abs/2112.09332} {Webgpt: Browser-assisted question-answering with human feedback}.
\newblock \emph{CoRR}, abs/2112.09332.

\bibitem[{Neelakantan et~al.(2022)Neelakantan, Xu, Puri, Radford, Han, Tworek, Yuan, Tezak, Kim, Hallacy, Heidecke, Shyam, Power, Nekoul, Sastry, Krueger, Schnurr, Such, Hsu, Thompson, Khan, Sherbakov, Jang, Welinder, and Weng}]{Neelakantan2022TextAC}
Arvind Neelakantan, Tao Xu, Raul Puri, Alec Radford, Jesse~Michael Han, Jerry Tworek, Qiming Yuan, Nikolas Tezak, Jong~Wook Kim, Chris Hallacy, Johannes Heidecke, Pranav Shyam, Boris Power, Tyna~Eloundou Nekoul, Girish Sastry, Gretchen Krueger, David Schnurr, Felipe~Petroski Such, Kenny Hsu, Madeleine Thompson, Tabarak Khan, Toki Sherbakov, Joanne Jang, Peter Welinder, and Lilian Weng. 2022.
\newblock \href {http://arxiv.org/abs/2201.10005} {Text and code embeddings by contrastive pre-training}.
\newblock \emph{CoRR}, abs/2201.10005.

\bibitem[{OpenAI(2023)}]{Achiam2023GPT4TR}
OpenAI. 2023.
\newblock \href {https://doi.org/10.48550/ARXIV.2303.08774} {{GPT-4} technical report}.
\newblock \emph{CoRR}, abs/2303.08774.

\bibitem[{Ouyang et~al.(2022)Ouyang, Wu, Jiang, Almeida, Wainwright, Mishkin, Zhang, Agarwal, Slama, Ray, Schulman, Hilton, Kelton, Miller, Simens, Askell, Welinder, Christiano, Leike, and Lowe}]{Ouyang2022TrainingLM}
Long Ouyang, Jeffrey Wu, Xu~Jiang, Diogo Almeida, Carroll~L. Wainwright, Pamela Mishkin, Chong Zhang, Sandhini Agarwal, Katarina Slama, Alex Ray, John Schulman, Jacob Hilton, Fraser Kelton, Luke Miller, Maddie Simens, Amanda Askell, Peter Welinder, Paul~F. Christiano, Jan Leike, and Ryan Lowe. 2022.
\newblock \href {http://papers.nips.cc/paper\_files/paper/2022/hash/b1efde53be364a73914f58805a001731-Abstract-Conference.html} {Training language models to follow instructions with human feedback}.
\newblock In \emph{Advances in Neural Information Processing Systems 35: Annual Conference on Neural Information Processing Systems 2022, NeurIPS 2022, New Orleans, LA, USA, November 28 - December 9, 2022}.

\bibitem[{Penedo et~al.(2023)Penedo, Malartic, Hesslow, Cojocaru, Alobeidli, Cappelli, Pannier, Almazrouei, and Launay}]{Penedo2023TheRD}
Guilherme Penedo, Quentin Malartic, Daniel Hesslow, Ruxandra Cojocaru, Hamza Alobeidli, Alessandro Cappelli, Baptiste Pannier, Ebtesam Almazrouei, and Julien Launay. 2023.
\newblock \href {http://papers.nips.cc/paper\_files/paper/2023/hash/fa3ed726cc5073b9c31e3e49a807789c-Abstract-Datasets\_and\_Benchmarks.html} {The refinedweb dataset for falcon {LLM:} outperforming curated corpora with web data only}.
\newblock In \emph{Advances in Neural Information Processing Systems 36: Annual Conference on Neural Information Processing Systems 2023, NeurIPS 2023, New Orleans, LA, USA, December 10 - 16, 2023}.

\bibitem[{Peng et~al.(2023)Peng, Galley, He, Cheng, Xie, Hu, Huang, Liden, Yu, Chen, and Gao}]{Peng2023CheckYF}
Baolin Peng, Michel Galley, Pengcheng He, Hao Cheng, Yujia Xie, Yu~Hu, Qiuyuan Huang, Lars Liden, Zhou Yu, Weizhu Chen, and Jianfeng Gao. 2023.
\newblock \href {https://doi.org/10.48550/ARXIV.2302.12813} {Check your facts and try again: Improving large language models with external knowledge and automated feedback}.
\newblock \emph{CoRR}, abs/2302.12813.

\bibitem[{Petroni et~al.(2020)Petroni, Lewis, Piktus, Rockt{\"{a}}schel, Wu, Miller, and Riedel}]{Petroni2020HowCA}
Fabio Petroni, Patrick S.~H. Lewis, Aleksandra Piktus, Tim Rockt{\"{a}}schel, Yuxiang Wu, Alexander~H. Miller, and Sebastian Riedel. 2020.
\newblock \href {https://doi.org/10.24432/C5201W} {How context affects language models' factual predictions}.
\newblock In \emph{Conference on Automated Knowledge Base Construction, {AKBC} 2020, Virtual, June 22-24, 2020}.

\bibitem[{Petroni et~al.(2021)Petroni, Piktus, Fan, Lewis, Yazdani, Cao, Thorne, Jernite, Karpukhin, Maillard, Plachouras, Rockt{\"{a}}schel, and Riedel}]{Petroni2020KILTAB}
Fabio Petroni, Aleksandra Piktus, Angela Fan, Patrick S.~H. Lewis, Majid Yazdani, Nicola~De Cao, James Thorne, Yacine Jernite, Vladimir Karpukhin, Jean Maillard, Vassilis Plachouras, Tim Rockt{\"{a}}schel, and Sebastian Riedel. 2021.
\newblock \href {https://doi.org/10.18653/V1/2021.NAACL-MAIN.200} {{KILT:} a benchmark for knowledge intensive language tasks}.
\newblock In \emph{Proceedings of the 2021 Conference of the North American Chapter of the Association for Computational Linguistics: Human Language Technologies, {NAACL-HLT} 2021, Online, June 6-11, 2021}, pages 2523--2544. Association for Computational Linguistics.

\bibitem[{Press et~al.(2023)Press, Zhang, Min, Schmidt, Smith, and Lewis}]{Press2022MeasuringAN}
Ofir Press, Muru Zhang, Sewon Min, Ludwig Schmidt, Noah~A. Smith, and Mike Lewis. 2023.
\newblock \href {https://doi.org/10.18653/V1/2023.FINDINGS-EMNLP.378} {Measuring and narrowing the compositionality gap in language models}.
\newblock In \emph{Findings of the Association for Computational Linguistics: {EMNLP} 2023, Singapore, December 6-10, 2023}, pages 5687--5711. Association for Computational Linguistics.

\bibitem[{Qin et~al.(2023{\natexlab{a}})Qin, Cai, Jin, Yan, Liang, Zhu, Lin, Han, Ding, Wang, Xie, Qi, Liu, Sun, and Zhou}]{Qin2023WebCPMIW}
Yujia Qin, Zihan Cai, Dian Jin, Lan Yan, Shihao Liang, Kunlun Zhu, Yankai Lin, Xu~Han, Ning Ding, Huadong Wang, Ruobing Xie, Fanchao Qi, Zhiyuan Liu, Maosong Sun, and Jie Zhou. 2023{\natexlab{a}}.
\newblock \href {https://doi.org/10.18653/V1/2023.ACL-LONG.499} {Webcpm: Interactive web search for chinese long-form question answering}.
\newblock In \emph{Proceedings of the 61st Annual Meeting of the Association for Computational Linguistics (Volume 1: Long Papers), {ACL} 2023, Toronto, Canada, July 9-14, 2023}, pages 8968--8988. Association for Computational Linguistics.

\bibitem[{Qin et~al.(2023{\natexlab{b}})Qin, Hu, Lin, Chen, Ding, Cui, Zeng, Huang, Xiao, Han, Fung, Su, Wang, Qian, Tian, Zhu, Liang, Shen, Xu, Zhang, Ye, Li, Tang, Yi, Zhu, Dai, Yan, Cong, Lu, Zhao, Huang, Yan, Han, Sun, Li, Phang, Yang, Wu, Ji, Liu, and Sun}]{Qin2023ToolLW}
Yujia Qin, Shengding Hu, Yankai Lin, Weize Chen, Ning Ding, Ganqu Cui, Zheni Zeng, Yufei Huang, Chaojun Xiao, Chi Han, Yi~Ren Fung, Yusheng Su, Huadong Wang, Cheng Qian, Runchu Tian, Kunlun Zhu, Shihao Liang, Xingyu Shen, Bokai Xu, Zhen Zhang, Yining Ye, Bowen Li, Ziwei Tang, Jing Yi, Yuzhang Zhu, Zhenning Dai, Lan Yan, Xin Cong, Yaxi Lu, Weilin Zhao, Yuxiang Huang, Junxi Yan, Xu~Han, Xian Sun, Dahai Li, Jason Phang, Cheng Yang, Tongshuang Wu, Heng Ji, Zhiyuan Liu, and Maosong Sun. 2023{\natexlab{b}}.
\newblock \href {https://doi.org/10.48550/ARXIV.2304.08354} {Tool learning with foundation models}.
\newblock \emph{CoRR}, abs/2304.08354.

\bibitem[{Ram et~al.(2023)Ram, Levine, Dalmedigos, Muhlgay, Shashua, Leyton{-}Brown, and Shoham}]{Ram2023InContextRL}
Ori Ram, Yoav Levine, Itay Dalmedigos, Dor Muhlgay, Amnon Shashua, Kevin Leyton{-}Brown, and Yoav Shoham. 2023.
\newblock \href {https://doi.org/10.48550/ARXIV.2302.00083} {In-context retrieval-augmented language models}.
\newblock \emph{CoRR}, abs/2302.00083.

\bibitem[{Robertson and Zaragoza(2009)}]{Robertson2009ThePR}
Stephen~E. Robertson and Hugo Zaragoza. 2009.
\newblock \href {https://doi.org/10.1561/1500000019} {The probabilistic relevance framework: {BM25} and beyond}.
\newblock \emph{Found. Trends Inf. Retr.}, 3(4):333--389.

\bibitem[{Rubin et~al.(2022)Rubin, Herzig, and Berant}]{Rubin2021LearningTR}
Ohad Rubin, Jonathan Herzig, and Jonathan Berant. 2022.
\newblock \href {https://doi.org/10.18653/V1/2022.NAACL-MAIN.191} {Learning to retrieve prompts for in-context learning}.
\newblock In \emph{Proceedings of the 2022 Conference of the North American Chapter of the Association for Computational Linguistics: Human Language Technologies, {NAACL} 2022, Seattle, WA, United States, July 10-15, 2022}, pages 2655--2671. Association for Computational Linguistics.

\bibitem[{Scao et~al.(2022)Scao, Fan, Akiki, Pavlick, Ilic, Hesslow, Castagn{\'{e}}, Luccioni, Yvon, Gall{\'{e}}, Tow, Rush, Biderman, Webson, Ammanamanchi, Wang, Sagot, Muennighoff, del Moral, Ruwase, Bawden, Bekman, McMillan{-}Major, Beltagy, Nguyen, Saulnier, Tan, Suarez, Sanh, Lauren{\c{c}}on, Jernite, Launay, Mitchell, Raffel, Gokaslan, Simhi, Soroa, Aji, Alfassy, Rogers, Nitzav, Xu, Mou, Emezue, Klamm, Leong, van Strien, Adelani, and et~al.}]{Scao2022BLOOMA1}
Teven~Le Scao, Angela Fan, Christopher Akiki, Ellie Pavlick, Suzana Ilic, Daniel Hesslow, Roman Castagn{\'{e}}, Alexandra~Sasha Luccioni, Fran{\c{c}}ois Yvon, Matthias Gall{\'{e}}, Jonathan Tow, Alexander~M. Rush, Stella Biderman, Albert Webson, Pawan~Sasanka Ammanamanchi, Thomas Wang, Beno{\^{\i}}t Sagot, Niklas Muennighoff, Albert~Villanova del Moral, Olatunji Ruwase, Rachel Bawden, Stas Bekman, Angelina McMillan{-}Major, Iz~Beltagy, Huu Nguyen, Lucile Saulnier, Samson Tan, Pedro~Ortiz Suarez, Victor Sanh, Hugo Lauren{\c{c}}on, Yacine Jernite, Julien Launay, Margaret Mitchell, Colin Raffel, Aaron Gokaslan, Adi Simhi, Aitor Soroa, Alham~Fikri Aji, Amit Alfassy, Anna Rogers, Ariel~Kreisberg Nitzav, Canwen Xu, Chenghao Mou, Chris Emezue, Christopher Klamm, Colin Leong, Daniel van Strien, David~Ifeoluwa Adelani, and et~al. 2022.
\newblock \href {https://doi.org/10.48550/ARXIV.2211.05100} {{BLOOM:} {A} 176b-parameter open-access multilingual language model}.
\newblock \emph{CoRR}, abs/2211.05100.

\bibitem[{Schick et~al.(2023)Schick, Dwivedi{-}Yu, Dess{\`{\i}}, Raileanu, Lomeli, Hambro, Zettlemoyer, Cancedda, and Scialom}]{Schick2023ToolformerLM}
Timo Schick, Jane Dwivedi{-}Yu, Roberto Dess{\`{\i}}, Roberta Raileanu, Maria Lomeli, Eric Hambro, Luke Zettlemoyer, Nicola Cancedda, and Thomas Scialom. 2023.
\newblock \href {http://papers.nips.cc/paper\_files/paper/2023/hash/d842425e4bf79ba039352da0f658a906-Abstract-Conference.html} {Toolformer: Language models can teach themselves to use tools}.
\newblock In \emph{Advances in Neural Information Processing Systems 36: Annual Conference on Neural Information Processing Systems 2023, NeurIPS 2023, New Orleans, LA, USA, December 10 - 16, 2023}.

\bibitem[{Shao et~al.(2023)Shao, Gong, Shen, Huang, Duan, and Chen}]{Shao2023EnhancingRL}
Zhihong Shao, Yeyun Gong, Yelong Shen, Minlie Huang, Nan Duan, and Weizhu Chen. 2023.
\newblock \href {https://doi.org/10.18653/V1/2023.FINDINGS-EMNLP.620} {Enhancing retrieval-augmented large language models with iterative retrieval-generation synergy}.
\newblock In \emph{Findings of the Association for Computational Linguistics: {EMNLP} 2023, Singapore, December 6-10, 2023}, pages 9248--9274. Association for Computational Linguistics.

\bibitem[{Shi et~al.(2023{\natexlab{a}})Shi, Chen, Misra, Scales, Dohan, Chi, Sch{\"{a}}rli, and Zhou}]{Shi2023LargeLM}
Freda Shi, Xinyun Chen, Kanishka Misra, Nathan Scales, David Dohan, Ed~H. Chi, Nathanael Sch{\"{a}}rli, and Denny Zhou. 2023{\natexlab{a}}.
\newblock \href {https://proceedings.mlr.press/v202/shi23a.html} {Large language models can be easily distracted by irrelevant context}.
\newblock In \emph{International Conference on Machine Learning, {ICML} 2023, 23-29 July 2023, Honolulu, Hawaii, {USA}}, volume 202 of \emph{Proceedings of Machine Learning Research}, pages 31210--31227. {PMLR}.

\bibitem[{Shi et~al.(2023{\natexlab{b}})Shi, Min, Yasunaga, Seo, James, Lewis, Zettlemoyer, and Yih}]{Shi2023REPLUGRB}
Weijia Shi, Sewon Min, Michihiro Yasunaga, Minjoon Seo, Rich James, Mike Lewis, Luke Zettlemoyer, and Wen{-}tau Yih. 2023{\natexlab{b}}.
\newblock \href {https://doi.org/10.48550/ARXIV.2301.12652} {{REPLUG:} retrieval-augmented black-box language models}.
\newblock \emph{CoRR}, abs/2301.12652.

\bibitem[{Shuster et~al.(2021)Shuster, Poff, Chen, Kiela, and Weston}]{Shuster2021RetrievalAR}
Kurt Shuster, Spencer Poff, Moya Chen, Douwe Kiela, and Jason Weston. 2021.
\newblock \href {https://doi.org/10.18653/V1/2021.FINDINGS-EMNLP.320} {Retrieval augmentation reduces hallucination in conversation}.
\newblock In \emph{Findings of the Association for Computational Linguistics: {EMNLP} 2021, Virtual Event / Punta Cana, Dominican Republic, 16-20 November, 2021}, pages 3784--3803. Association for Computational Linguistics.

\bibitem[{Stelmakh et~al.(2022)Stelmakh, Luan, Dhingra, and Chang}]{Stelmakh2022ASQAFQ}
Ivan Stelmakh, Yi~Luan, Bhuwan Dhingra, and Ming{-}Wei Chang. 2022.
\newblock \href {https://doi.org/10.18653/V1/2022.EMNLP-MAIN.566} {{ASQA:} factoid questions meet long-form answers}.
\newblock In \emph{Proceedings of the 2022 Conference on Empirical Methods in Natural Language Processing, {EMNLP} 2022, Abu Dhabi, United Arab Emirates, December 7-11, 2022}, pages 8273--8288. Association for Computational Linguistics.

\bibitem[{Tahami et~al.(2020)Tahami, Ghajar, and Shakery}]{DBLP:conf/sigir/TahamiGS20}
Amir~Vakili Tahami, Kamyar Ghajar, and Azadeh Shakery. 2020.
\newblock \href {https://doi.org/10.1145/3397271.3401296} {Distilling knowledge for fast retrieval-based chat-bots}.
\newblock In \emph{Proceedings of the 43rd International {ACM} {SIGIR} conference on research and development in Information Retrieval, {SIGIR} 2020, Virtual Event, China, July 25-30, 2020}, pages 2081--2084. {ACM}.

\bibitem[{Tao et~al.(2019)Tao, Wu, Xu, Hu, Zhao, and Yan}]{DBLP:conf/wsdm/TaoWXHZY19}
Chongyang Tao, Wei Wu, Can Xu, Wenpeng Hu, Dongyan Zhao, and Rui Yan. 2019.
\newblock \href {https://doi.org/10.1145/3289600.3290985} {Multi-representation fusion network for multi-turn response selection in retrieval-based chatbots}.
\newblock In \emph{Proceedings of the Twelfth {ACM} International Conference on Web Search and Data Mining, {WSDM} 2019, Melbourne, VIC, Australia, February 11-15, 2019}, pages 267--275. {ACM}.

\bibitem[{Touvron et~al.(2023)Touvron, Martin, Stone, Albert, Almahairi, Babaei, Bashlykov, Batra, Bhargava, Bhosale, Bikel, Blecher, Canton{-}Ferrer, Chen, Cucurull, Esiobu, Fernandes, Fu, Fu, Fuller, Gao, Goswami, Goyal, Hartshorn, Hosseini, Hou, Inan, Kardas, Kerkez, Khabsa, Kloumann, Korenev, Koura, Lachaux, Lavril, Lee, Liskovich, Lu, Mao, Martinet, Mihaylov, Mishra, Molybog, Nie, Poulton, Reizenstein, Rungta, Saladi, Schelten, Silva, Smith, Subramanian, Tan, Tang, Taylor, Williams, Kuan, Xu, Yan, Zarov, Zhang, Fan, Kambadur, Narang, Rodriguez, Stojnic, Edunov, and Scialom}]{Touvron2023Llama2O}
Hugo Touvron, Louis Martin, Kevin Stone, Peter Albert, Amjad Almahairi, Yasmine Babaei, Nikolay Bashlykov, Soumya Batra, Prajjwal Bhargava, Shruti Bhosale, Dan Bikel, Lukas Blecher, Cristian Canton{-}Ferrer, Moya Chen, Guillem Cucurull, David Esiobu, Jude Fernandes, Jeremy Fu, Wenyin Fu, Brian Fuller, Cynthia Gao, Vedanuj Goswami, Naman Goyal, Anthony Hartshorn, Saghar Hosseini, Rui Hou, Hakan Inan, Marcin Kardas, Viktor Kerkez, Madian Khabsa, Isabel Kloumann, Artem Korenev, Punit~Singh Koura, Marie{-}Anne Lachaux, Thibaut Lavril, Jenya Lee, Diana Liskovich, Yinghai Lu, Yuning Mao, Xavier Martinet, Todor Mihaylov, Pushkar Mishra, Igor Molybog, Yixin Nie, Andrew Poulton, Jeremy Reizenstein, Rashi Rungta, Kalyan Saladi, Alan Schelten, Ruan Silva, Eric~Michael Smith, Ranjan Subramanian, Xiaoqing~Ellen Tan, Binh Tang, Ross Taylor, Adina Williams, Jian~Xiang Kuan, Puxin Xu, Zheng Yan, Iliyan Zarov, Yuchen Zhang, Angela Fan, Melanie Kambadur, Sharan Narang, Aur{\'{e}}lien Rodriguez, Robert Stojnic, Sergey Edunov,
  and Thomas Scialom. 2023.
\newblock \href {https://doi.org/10.48550/ARXIV.2307.09288} {Llama 2: Open foundation and fine-tuned chat models}.
\newblock \emph{CoRR}, abs/2307.09288.

\bibitem[{Trivedi et~al.(2022)Trivedi, Balasubramanian, Khot, and Sabharwal}]{Trivedi2021MM}
Harsh Trivedi, Niranjan Balasubramanian, Tushar Khot, and Ashish Sabharwal. 2022.
\newblock \href {https://doi.org/10.1162/TACL\_A\_00475} {Musique: Multihop questions via single-hop question composition}.
\newblock \emph{Trans. Assoc. Comput. Linguistics}, 10:539--554.

\bibitem[{Wang et~al.(2022)Wang, Yang, Huang, Jiao, Yang, Jiang, Majumder, and Wei}]{Wang2022TextEB}
Liang Wang, Nan Yang, Xiaolong Huang, Binxing Jiao, Linjun Yang, Daxin Jiang, Rangan Majumder, and Furu Wei. 2022.
\newblock \href {https://doi.org/10.48550/ARXIV.2212.03533} {Text embeddings by weakly-supervised contrastive pre-training}.
\newblock \emph{CoRR}, abs/2212.03533.

\bibitem[{Wang et~al.(2023{\natexlab{a}})Wang, Yang, and Wei}]{Wang2023Query2docQE}
Liang Wang, Nan Yang, and Furu Wei. 2023{\natexlab{a}}.
\newblock \href {https://doi.org/10.18653/V1/2023.EMNLP-MAIN.585} {Query2doc: Query expansion with large language models}.
\newblock In \emph{Proceedings of the 2023 Conference on Empirical Methods in Natural Language Processing, {EMNLP} 2023, Singapore, December 6-10, 2023}, pages 9414--9423. Association for Computational Linguistics.

\bibitem[{Wang et~al.(2023{\natexlab{b}})Wang, Li, Sun, and Liu}]{Wang2023SelfKnowledgeGR}
Yile Wang, Peng Li, Maosong Sun, and Yang Liu. 2023{\natexlab{b}}.
\newblock \href {https://doi.org/10.18653/V1/2023.FINDINGS-EMNLP.691} {Self-knowledge guided retrieval augmentation for large language models}.
\newblock In \emph{Findings of the Association for Computational Linguistics: {EMNLP} 2023, Singapore, December 6-10, 2023}, pages 10303--10315. Association for Computational Linguistics.

\bibitem[{Wei et~al.(2022)Wei, Wang, Schuurmans, Bosma, Ichter, Xia, Chi, Le, and Zhou}]{Wei2022ChainOT}
Jason Wei, Xuezhi Wang, Dale Schuurmans, Maarten Bosma, Brian Ichter, Fei Xia, Ed~H. Chi, Quoc~V. Le, and Denny Zhou. 2022.
\newblock \href {http://papers.nips.cc/paper\_files/paper/2022/hash/9d5609613524ecf4f15af0f7b31abca4-Abstract-Conference.html} {Chain-of-thought prompting elicits reasoning in large language models}.
\newblock In \emph{Advances in Neural Information Processing Systems 35: Annual Conference on Neural Information Processing Systems 2022, NeurIPS 2022, New Orleans, LA, USA, November 28 - December 9, 2022}.

\bibitem[{White(2023)}]{White2023NavigatingCS}
Ryen~W. White. 2023.
\newblock \href {https://doi.org/10.48550/ARXIV.2311.01235} {Navigating complex search tasks with {AI} copilots}.
\newblock \emph{CoRR}, abs/2311.01235.

\bibitem[{Xu et~al.(2023)Xu, Shi, and Choi}]{Xu2023RECOMPIR}
Fangyuan Xu, Weijia Shi, and Eunsol Choi. 2023.
\newblock \href {https://doi.org/10.48550/ARXIV.2310.04408} {{RECOMP:} improving retrieval-augmented lms with compression and selective augmentation}.
\newblock \emph{CoRR}, abs/2310.04408.

\bibitem[{Yang et~al.(2023)Yang, Xiao, Wang, Zhang, Bian, Yin, Lv, Pan, Wang, Yan, Yang, Deng, Wang, Liu, Ai, Dong, Zhao, Xu, Sun, Zhang, Liu, Ji, Xie, Dai, Fang, Su, Song, Liu, Ru, Ma, Wang, Liu, Lin, Nie, Guo, Sun, Zhang, Li, Li, Cheng, Chen, Zeng, Wang, Chen, Men, Yu, Pan, Shen, Wang, Li, Jiang, Gao, Zhang, Zhou, and Wu}]{Yang2023Baichuan2O}
Aiyuan Yang, Bin Xiao, Bingning Wang, Borong Zhang, Ce~Bian, Chao Yin, Chenxu Lv, Da~Pan, Dian Wang, Dong Yan, Fan Yang, Fei Deng, Feng Wang, Feng Liu, Guangwei Ai, Guosheng Dong, Haizhou Zhao, Hang Xu, Haoze Sun, Hongda Zhang, Hui Liu, Jiaming Ji, Jian Xie, Juntao Dai, Kun Fang, Lei Su, Liang Song, Lifeng Liu, Liyun Ru, Luyao Ma, Mang Wang, Mickel Liu, MingAn Lin, Nuolan Nie, Peidong Guo, Ruiyang Sun, Tao Zhang, Tianpeng Li, Tianyu Li, Wei Cheng, Weipeng Chen, Xiangrong Zeng, Xiaochuan Wang, Xiaoxi Chen, Xin Men, Xin Yu, Xuehai Pan, Yanjun Shen, Yiding Wang, Yiyu Li, Youxin Jiang, Yuchen Gao, Yupeng Zhang, Zenan Zhou, and Zhiying Wu. 2023.
\newblock \href {https://doi.org/10.48550/ARXIV.2309.10305} {Baichuan 2: Open large-scale language models}.
\newblock \emph{CoRR}, abs/2309.10305.

\bibitem[{Yao et~al.(2023)Yao, Zhao, Yu, Du, Shafran, Narasimhan, and Cao}]{Yao2022ReActSR}
Shunyu Yao, Jeffrey Zhao, Dian Yu, Nan Du, Izhak Shafran, Karthik~R. Narasimhan, and Yuan Cao. 2023.
\newblock \href {https://openreview.net/pdf?id=WE\_vluYUL-X} {React: Synergizing reasoning and acting in language models}.
\newblock In \emph{The Eleventh International Conference on Learning Representations, {ICLR} 2023, Kigali, Rwanda, May 1-5, 2023}. OpenReview.net.

\bibitem[{Yoran et~al.(2023)Yoran, Wolfson, Ram, and Berant}]{Yoran2023MakingRL}
Ori Yoran, Tomer Wolfson, Ori Ram, and Jonathan Berant. 2023.
\newblock \href {https://doi.org/10.48550/ARXIV.2310.01558} {Making retrieval-augmented language models robust to irrelevant context}.
\newblock \emph{CoRR}, abs/2310.01558.

\bibitem[{Yu et~al.(2023{\natexlab{a}})Yu, Zhang, Pan, Ma, Wang, and Yu}]{Yu2023ChainofNoteER}
Wenhao Yu, Hongming Zhang, Xiaoman Pan, Kaixin Ma, Hongwei Wang, and Dong Yu. 2023{\natexlab{a}}.
\newblock \href {https://doi.org/10.48550/ARXIV.2311.09210} {Chain-of-note: Enhancing robustness in retrieval-augmented language models}.
\newblock \emph{CoRR}, abs/2311.09210.

\bibitem[{Yu et~al.(2023{\natexlab{b}})Yu, Zhang, Liang, Jiang, and Sabharwal}]{Yu2023ImprovingLM}
Wenhao Yu, Zhihan Zhang, Zhenwen Liang, Meng Jiang, and Ashish Sabharwal. 2023{\natexlab{b}}.
\newblock \href {https://doi.org/10.48550/ARXIV.2305.14002} {Improving language models via plug-and-play retrieval feedback}.
\newblock \emph{CoRR}, abs/2305.14002.

\bibitem[{Zhai et~al.(2023)Zhai, Tong, Li, Cai, Qu, Lee, and Ma}]{Zhai2023InvestigatingTC}
Yuexiang Zhai, Shengbang Tong, Xiao Li, Mu~Cai, Qing Qu, Yong~Jae Lee, and Yi~Ma. 2023.
\newblock \href {https://doi.org/10.48550/ARXIV.2309.10313} {Investigating the catastrophic forgetting in multimodal large language models}.
\newblock \emph{CoRR}, abs/2309.10313.

\bibitem[{Zhang et~al.(2024)Zhang, Zeng, Wang, and Lu}]{Zhang2024TinyLlamaAO}
Peiyuan Zhang, Guangtao Zeng, Tianduo Wang, and Wei Lu. 2024.
\newblock \href {https://doi.org/10.48550/ARXIV.2401.02385} {Tinyllama: An open-source small language model}.
\newblock \emph{CoRR}, abs/2401.02385.

\bibitem[{Zhou et~al.(2021)Zhou, Neubig, Gu, Diab, Guzm{\'{a}}n, Zettlemoyer, and Ghazvininejad}]{Zhou2020DetectingHC}
Chunting Zhou, Graham Neubig, Jiatao Gu, Mona~T. Diab, Francisco Guzm{\'{a}}n, Luke Zettlemoyer, and Marjan Ghazvininejad. 2021.
\newblock \href {https://doi.org/10.18653/V1/2021.FINDINGS-ACL.120} {Detecting hallucinated content in conditional neural sequence generation}.
\newblock In \emph{Findings of the Association for Computational Linguistics: {ACL/IJCNLP} 2021, Online Event, August 1-6, 2021}, volume {ACL/IJCNLP} 2021 of \emph{Findings of {ACL}}, pages 1393--1404. Association for Computational Linguistics.

\bibitem[{Zhu et~al.(2023)Zhu, Yuan, Wang, Liu, Liu, Deng, Dou, and Wen}]{Zhu2023LargeLM}
Yutao Zhu, Huaying Yuan, Shuting Wang, Jiongnan Liu, Wenhan Liu, Chenlong Deng, Zhicheng Dou, and Ji{-}Rong Wen. 2023.
\newblock \href {https://doi.org/10.48550/ARXIV.2308.07107} {Large language models for information retrieval: {A} survey}.
\newblock \emph{CoRR}, abs/2308.07107.

\bibitem[{Zhu et~al.(2024)Zhu, Zhang, Zhang, Chen, Xie, Dou, Liu, and Wen}]{inters}
Yutao Zhu, Peitian Zhang, Chenghao Zhang, Yifei Chen, Binyu Xie, Zhicheng Dou, Zheng Liu, and Ji{-}Rong Wen. 2024.
\newblock \href {https://doi.org/10.48550/ARXIV.2401.06532} {{INTERS:} unlocking the power of large language models in search with instruction tuning}.
\newblock \emph{CoRR}, abs/2401.06532.

\end{thebibliography}
\end{document}